\documentclass[10pt,twocolumn,letterpaper]{article}

\usepackage[pagenumbers]{cvpr} 
\usepackage[dvipsnames]{xcolor}
\newcommand{\red}[1]{{\color{red}#1}}

\usepackage[title]{appendix}

\usepackage{times}
\usepackage{epsfig}
\usepackage{graphicx}
\usepackage{amsmath}
\usepackage{amssymb}
\usepackage{booktabs}
\usepackage{comment}
\usepackage{array}
\usepackage[accsupp]{axessibility} 

\definecolor{cvprblue}{rgb}{0.21,0.49,0.74}
\usepackage[pagebackref,breaklinks,colorlinks,citecolor=cvprblue]{hyperref}

\newcolumntype{L}[1]{>{\raggedright\arraybackslash}m{#1}}
\newcolumntype{C}[1]{>{\centering\arraybackslash}m{#1}}
\newcolumntype{R}[1]{>{\raggedleft\arraybackslash}m{#1}}
\newcolumntype{+}{>{\global\let\currentrowstyle\relax}}
\newcolumntype{^}{>{\currentrowstyle}}

\newcommand{\RomNum}[1]{\MakeUppercase{\romannumeral #1}}

\newcommand{\bfp}{\ensuremath{{\mathbf{p}}}}

\newcommand{\bfc}{\ensuremath{{\mathbf{c}}}}
\newcommand{\bfC}{\ensuremath{{\mathbf{C}}}}

\def\paperID{3743}
\def\confName{CVPR}
\def\confYear{2024}

\begin{document}
\title{Semantic Line Combination Detector}

\author{Jinwon Ko\\
Korea University\\
{\tt\small jwko@mcl.korea.ac.kr}
\and
Dongkwon Jin\\
Korea University\\
{\tt\small dongkwonjin@mcl.korea.ac.kr}
\and
Chang-Su Kim\thanks{Corresponding author.}\\
Korea University\\
{\tt\small changsukim@korea.ac.kr}}

\maketitle
\begin{abstract}
   A novel algorithm, called semantic line combination detector (SLCD), to find an optimal combination of semantic lines is proposed in this paper. It processes all lines in each line combination at once to assess the overall harmony of the lines. First, we generate various line combinations from reliable lines. Second, we estimate the score of each line combination and determine the best one. Experimental results demonstrate that the proposed SLCD outperforms existing semantic line detectors on various datasets. Moreover, it is shown that SLCD can be applied effectively to three vision tasks of vanishing point detection, symmetry axis detection, and composition-based image retrieval. Our codes are available at \href{https://github.com/Jinwon-Ko/SLCD}{https://github.com/Jinwon-Ko/SLCD}.
\end{abstract}
\vspace{-0.4cm}
\section{Introduction}
\vspace{-0.1cm}
A \textit{semantic line} \cite{lee2017semantic} is defined as a meaningful line separating distinct semantic regions in an image. Besides this unary definition, multiple semantic lines in an image are supposed to convey the global scene structure properly \cite{jin2021harmonious}. It is challenging to detect such semantic lines because they are often implied by complicated region boundaries. Moreover, they should represent the image composition optimally by dividing it into semantic regions harmoniously.

\begin{figure}
\begin{center}
\includegraphics[width=1\linewidth]{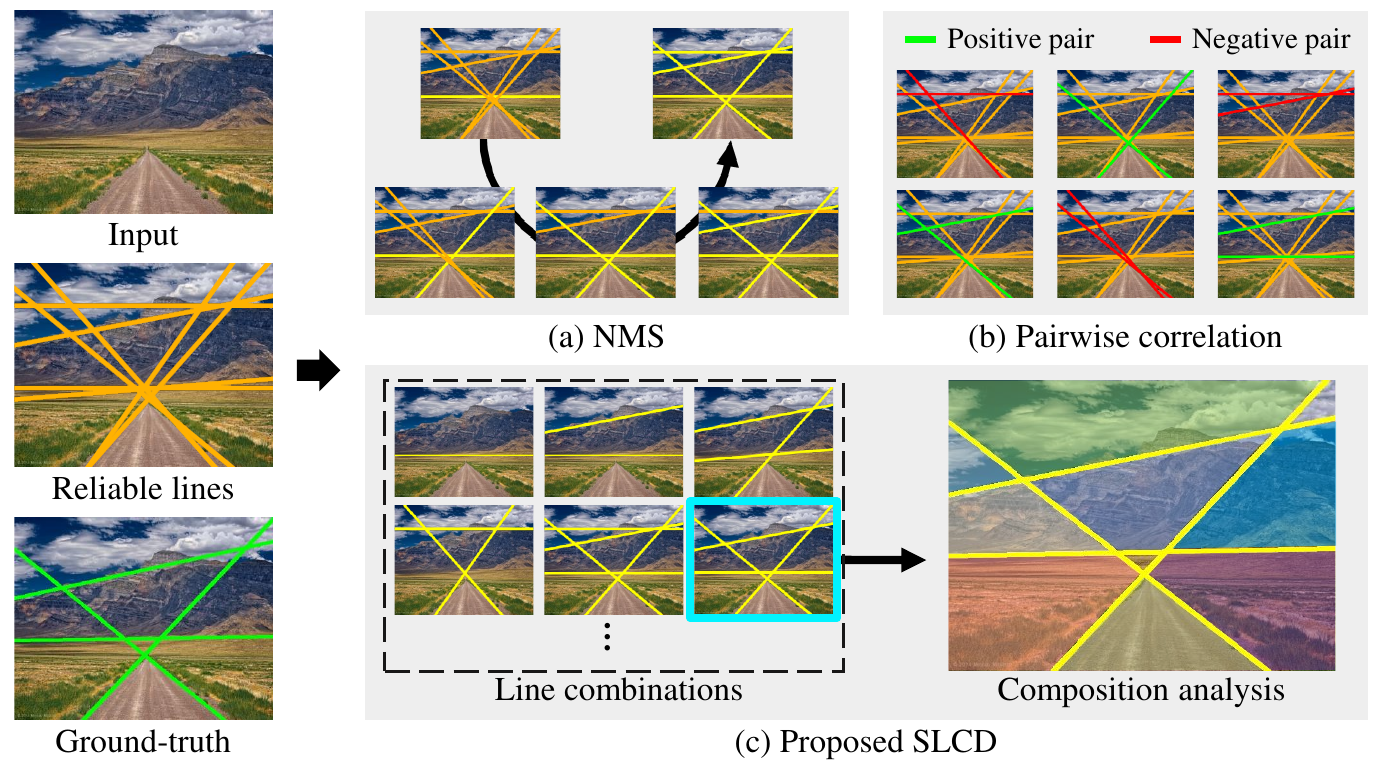}
\end{center}
\vspace{-0.4cm}
   \caption{After selecting reliable line candidates, there are two existing approaches to semantic line detection. The first approach in (a) focuses on locating a line near a region boundary and eliminating overlapping lines. However, a redundant line still remains, since this approach does not consider how well a group of detected lines represents the layout of a scene. The second approach in (b) takes into account only the pairwise correlation between two lines, so it may fail to assess the overall harmony of more than two semantic lines. In contrast, in (c), the proposed SLCD generates a number of line combinations, analyzes all lines in each combination at once, and then finds the most harmonious combination that conveys the global scene composition optimally.}
\label{fig:Intro}
\vspace{-0.2cm}
\end{figure}

Semantic lines are essential elements in many vision applications. For example,  a horizon line \cite{workman2016horizon,zhai2016detecting,zhu2020single,kluger2020temporally}, which is a specific type of semantic line, can be exploited to adjust the levelness of an image \cite{lee2017semantic,wang2023spatial}. A reflection symmetry axis \cite{funk2017,cicconet2017_nyu,elawady2017,loy2006detecting,cicconet2017finding}, which is another type of semantic line, provides visual cues for object recognition and pattern analysis. Vanishing points, conveying depth impression in images, can be estimated by detecting dominant parallel semantic lines in the 3D world \cite{zhou2017detecting,jin2020semantic,bazin2012globally}. In autonomous driving systems \cite{tabelini2021CVPR,jin2021harmonious,zheng2022clrnet}, boundaries of road lanes can be also described by semantic lines.

\begin{figure*}[t]
\begin{center}
\includegraphics[width=1\linewidth]{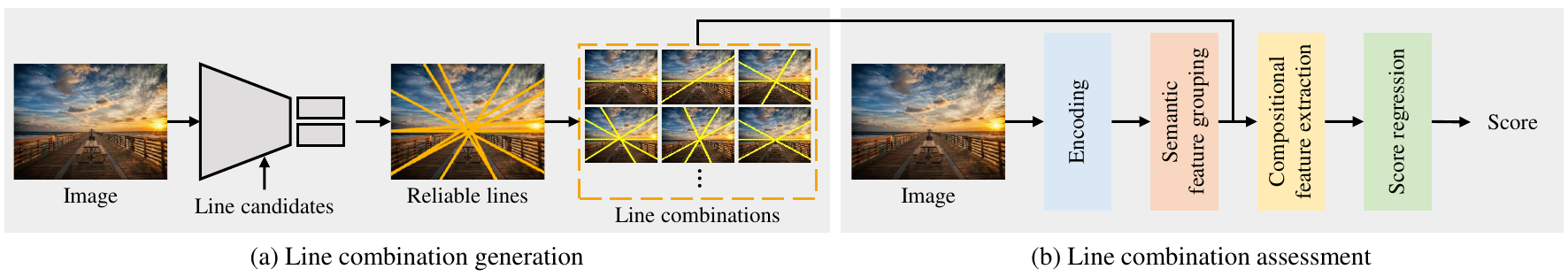}
\end{center}
\vspace{-0.4cm}
   \caption{Overview of the proposed SLCD algorithm.}
\label{fig:Overview}
\vspace{-0.1cm}
\end{figure*}

Recently, several attempts \cite{lee2017semantic,jin2020semantic,han2020eccv,zhao2021deep,jin2021harmonious} have been made to detect semantic lines. These techniques perform line detection and refinement sequentially. At the detection stage, they extract deep line features to classify and regress each line candidate. At the refinement stage, reliable semantic lines are determined by removing redundant lines. Specifically, to refine line candidates, non-maximum suppression (NMS) is performed in \cite{lee2017semantic,han2020eccv,jin2020semantic}, as illustrated in Figure~\ref{fig:Intro}(a). Lee \etal \cite{lee2017semantic} iteratively select the reliable line near boundary pixels and remove overlapping lines with the selected one. Han \etal \cite{han2020eccv} simplify the NMS process by adopting a Hough line space. Jin \etal \cite{jin2020semantic} process each candidate through comparative ranking and matching. These techniques, however, do not consider the overall harmony among detected lines.
To cope with this issue, Jin \etal \cite{jin2021harmonious} estimate the relation score for every pair of detected lines and then decide harmonious semantic lines via graph optimization. But, they may yield sub-optimal results, since only the pairwise relationships between lines are exploited, as in Figure~\ref{fig:Intro}(b).

In this paper, we propose a novel algorithm, called semantic line combination detector (SLCD), to find an optimal group of semantic lines. It processes all lines in a line group (or combination) simultaneously, instead of analyzing each pair of lines, to estimate the overall harmony, as in Figure~\ref{fig:Intro}(c). Figure~\ref{fig:Overview} shows an overview of the proposed SLCD. First, we select reliable lines from line candidates and then generate a number of line combinations. Second, we score all line combinations and determine the combination with the highest score as the optimal group of semantic lines. To this end, we design two novel modules for semantic feature grouping and compositional feature extraction. We also introduce a novel loss function to guide the feature grouping module. Experimental results demonstrate that SLCD can detect semantic lines reliably on existing datasets and a new dataset, called compositionally diverse lines (CDL). Moreover, it is shown that SLCD can be used effectively in various applications.

This work has the following major contributions:
\begin{itemize}
\item SLCD finds an optimal combination of semantic lines by processing all lines in a line combination at once.
\item We construct the CDL dataset containing compositionally diverse images with implied lines. It will be made publicly available.\footnote{CDL is available at \href{https://github.com/Jinwon-Ko/SLCD}{https://github.com/Jinwon-Ko/SLCD}.}
\item SLCD outperforms conventional detectors on most datasets. Also, its effectiveness is demonstrated in three applications: vanishing point detection, symmetry axis detection, and composition-based image retrieval.
\end{itemize}

\vspace{-0.1cm}
\section{Related Work}
Semantic lines, located near the boundaries of different semantic regions, outline the layout and composition of an image. They play an important role in various vision applications. Horizons \cite{workman2016horizon,zhai2016detecting,zhu2020single,kluger2020temporally} are a specific type of semantic lines, which can be applied to adjust the levelness of images and improve their aesthetics \cite{lee2017semantic,wang2023spatial}. In \cite{zhou2017detecting,jin2020semantic,bazin2012globally}, dominant parallel lines in the 3D world are detected to estimate vanishing points, which convey depth impression on 2D images. Also, in \cite{loy2006detecting,funk2017,cicconet2017_nyu,elawady2017}, reflection symmetry axes are identified to analyze the shapes of objects or patterns. These types of lines can be regarded as highly implied semantic lines. In \cite{tabelini2021CVPR,jin2021harmonious,zheng2022clrnet}, straight lanes are detected to aid in vehicle maneuvers in road environments. Furthermore, semantic lines are essential visual cues in photographic composition \cite{liu2010optimizing,lee2018photographic}. They direct viewers' attention and help to compose a visually balanced image.

\begin{figure*}[t]
\begin{center}
\includegraphics[width=1\linewidth]{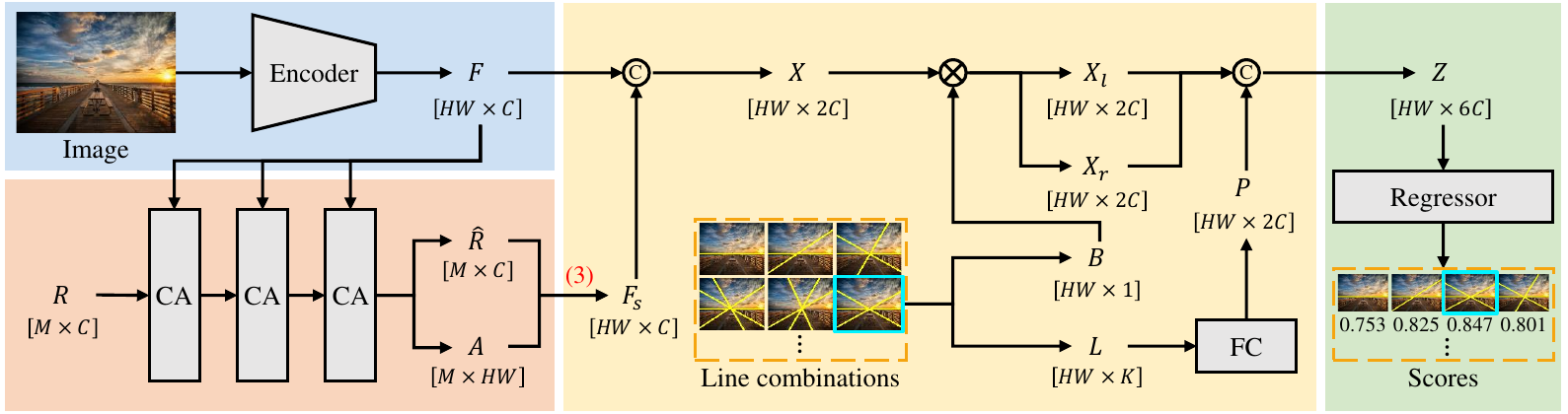}
\end{center}
\vspace{-0.4cm}
   \caption{The architecture of SLCD. CA and FC denote cross-attention and fully connected layers, respectively. The blue, red, yellow, and green boxes indicate encoding, semantic feature grouping, compositional feature extraction, and score regression, respectively.}
\label{fig:SLCD}
\end{figure*}

Several semantic line detectors \cite{lee2017semantic,han2020eccv,jin2020semantic,jin2021harmonious} have been proposed. They perform in two stages: line detection and refinement. At the detection stage, deep line features are extracted to classify and regress each line candidate. At the refinement stage, reliable semantic lines are determined by removing irrelevant candidates, based on NMS \cite{lee2017semantic,han2020eccv,jin2020semantic} or graph optimization \cite{jin2021harmonious}. More specifically, Lee \etal \cite{lee2017semantic} detect reliable lines with classification probabilities higher than a threshold. They then iteratively select semantic lines and remove overlapping lines with the selected one, by employing the edge detector in \cite{xie2015holistically}. Han \etal \cite{han2020eccv} predict the probability of each candidate in a Hough parametric space. They simplify NMS by computing the centroids of connected components in the Hough space. Jin \etal \cite{jin2020semantic} design two comparators to estimate the priority and similarity between two lines. Then, they determine the most reliable lines and eliminate redundant ones alternately through pairwise comparisons. However, these techniques \cite{lee2017semantic,han2020eccv,jin2020semantic} do not consider how well a group of detected lines represents the global scene structure. To address this issue, Jin \etal \cite{jin2021harmonious} analyze the pairwise harmony of detected lines. They first estimate a harmony score for a pair of detected lines. They then construct a complete graph and determine harmonious semantic lines by finding a maximal weight clique. Their method, however, may yield sub-optimal results because it considers only pairwise relationships between lines. On the contrary, the proposed SLCD finds an optimal combination of semantic lines by analyzing all lines in each combination simultaneously.

\vspace{-0.1cm}
\section{Proposed Algorithm}
We propose a novel algorithm, called SLCD, to detect an optimal combination of semantic lines, an overview of which is in Figure~\ref{fig:Overview}. First, we select $K$ reliable lines from line candidates and then generate a number of line combinations. Second, we score all the line combinations and determine the combination with the highest score as the optimal group of semantic lines.
\subsection{Generating Line Combinations}\label{subsec:comb}
\noindent {\bf Initializing line candidates:} We generate line candidates, which are end-to-end straight lines in an image. Each line candidate is parameterized by polar coordinates in the Hough space \cite{han2020eccv}. By quantizing the coordinates uniformly, we obtain $N$ line candidates. The default $N$ is 1024.

\begin{figure}
\begin{center}
\includegraphics[width=1\linewidth]{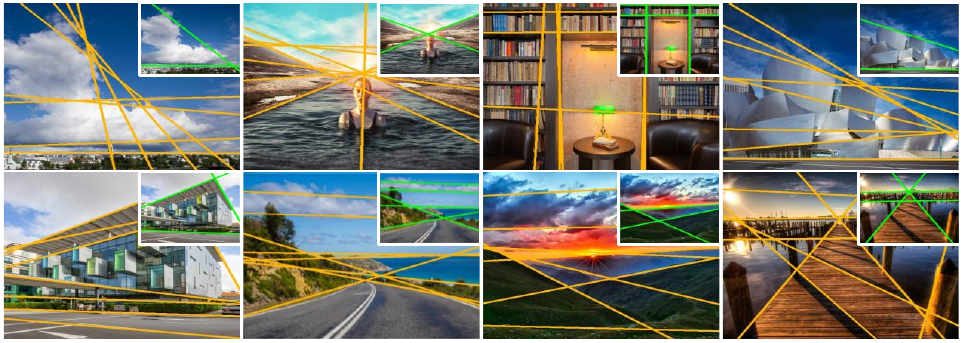}
\end{center}
\vspace{-0.4cm}
   \caption{From a set of line candidates, the line detector selects $K$ reliable lines, depicted in orange. A high recall rate is achieved because a sufficient number of reliable lines are selected through NMS. Ground-truth semantic lines in green are in the insets.}
\label{fig:filtering}
\end{figure}

\vspace*{0.1cm}
\noindent {\bf Filtering line candidates:} Given the $N$ line candidates, there are $2^N$ possible line combinations in total. Since this number of combinations is unmanageable, we filter out irrelevant lines to maintain $K$ reliable ones only. To this end, we design a simple line detector by modifying S-Net in \cite{jin2021harmonious}. Given an image, the detector obtains a convolutional feature map and extracts line features by aggregating the features of pixels along each line candidate. Then, it computes the classification probability and regression offset of each candidate. We select the most reliable candidate with the highest probability and remove overlapping lines with the selected one. We iterate this process $K$ times to determine $K$ reliable lines. The default $K$ is 8. Figure~\ref{fig:filtering} shows examples of selected reliable lines. Note that, in this filtering, false negatives occur rarely because the $K$ lines are selected via NMS even though their classification probabilities are not high. The architecture of the modified S-Net is described in detail in the appendix (Section A.1).

\vspace*{0.1cm}
\noindent {\bf Generating line combinations:} From the $K$ reliable lines, we generate all $2^K$ line combinations.

\subsection{Assessing Line Combinations}

When a line combination divides an image insufficiently or over-segments it into unnecessary parts, it does not describe the overall structure of the scene properly. On the contrary, an optimal combination of semantic lines should convey the image composition reliably and efficiently (\ie with a small number of lines). To find the best line combination, we develop the semantic line combination detector (SLCD). Figure~\ref{fig:SLCD} shows the structure of SLCD, which performs encoding, semantic feature grouping, compositional feature extraction, and score regression. The detailed architecture of SLCD is described in the appendix (Section A.2).

\vspace*{0.1cm}
\noindent {\bf Encoding:}
Given an image, we extract multi-scale feature maps of ResNet50 \cite{he2016deep}. Then, we match their spatial resolutions via bilinear interpolation and concatenate them in the channel dimension. We then squeeze the channels using 2D convolutional layers to obtain an aggregated feature map $F \in \mathbb{R}^{HW \times C}$, where $H$, $W$, and $C$ are the feature height, the feature width, and the number of channels.

\vspace*{0.1cm}
\noindent {\bf Semantic feature grouping:}
Semantic lines are located near the boundaries between distinct regions. For their reliable detection, it is desirable to separate different regional parts more clearly. We hence attempt to group pixel features into multiple regions through cross-attention \cite{carion2020end, yu2022cmt,yu2022kMaX,piccinelli2023idisc}. We employ three cross-attention modules.

Let $R \in \mathbb{R}^{M\times C}$ be a learnable matrix representing $M$ regions, called region query matrix. Then, we convert $R$ into queries and $F$ into keys and values by
\begin{equation}
    R_q = RU_{q}, \,\,\, F_k = (F + S) U_{k}, \,\,\, F_v = (F + S) U_{v},
\end{equation}
where $S \in \mathbb{R}^{HW \times C}$ denotes the sinusoidal positional encoding \cite{vaswani2017attention}, and $U_{q}, U_{k}, U_{v} \in \mathbb{R}^{C \times C}$ are projection matrices for queries, keys, and values. We then obtain an updated region query matrix $\hat{R}$ by
\begin{equation}
    A = \underset{M}{\operatorname{softmax}}(R_q F_k^{\rm T}/\tau), \,\,\, \hat{R} = AF_v + R,
    \label{eq:region_partitioning}
\end{equation}
where $A \in \mathbb{R}^{M \times HW}$ is the attention matrix with a scaling factor $\tau$, and the letter `$M$' means the softmax operation is done in the column direction, as in \cite{yu2022cmt,yu2022kMaX,piccinelli2023idisc}. Then, in the last cross-attention, we generate a semantic feature map $F_s \in \mathbb{R}^{HW \times C}$ by
\begin{equation}
    F_s = A^{\rm T}\hat{R}.
\label{eq:regional}
\end{equation}
Finally, we channel-wise concatenate $F$ and $F_s$ to obtain a combined feature map $X \in \mathbb{R}^{HW \times 2C}$.

\begin{figure}
\begin{center}
\includegraphics[width=1\linewidth]{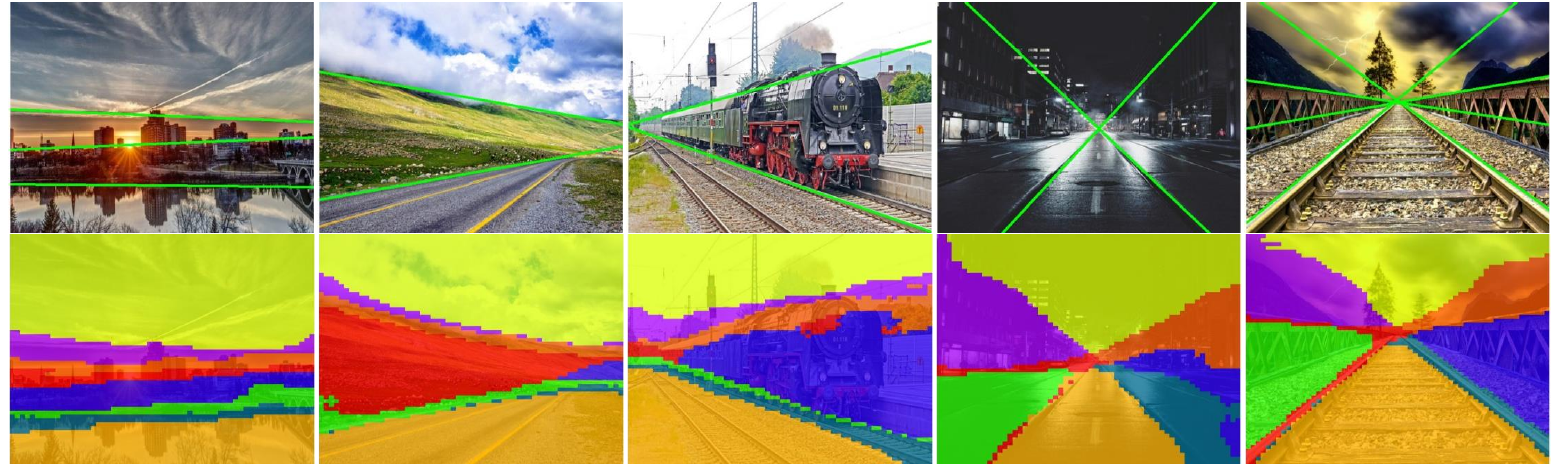}
\end{center}
\vspace{-0.4cm}
   \caption{Visualization of semantic feature grouping results. The top row shows input images with ground-truth lines. The bottom one presents the membership maps, representing the semantic region that each pixel belongs to. }
\label{fig:attn_map}
\vspace{-0.3cm}
\end{figure}

Note that, in (\ref{eq:region_partitioning}), the softmax function is applied along the query axis. Thus, in $A=[\bfp_{1}, \bfp_{2}, \dots, \bfp_{HW}]$, each column $\bfp_{i} \in \mathbb{R}^{M}$ is a probability vector. Specifically, the $m$th element $p_{i}^{m}$ in $\bfp_{i}$ is the probability that pixel $i$ belongs to the $m$th region query. Figure~\ref{fig:attn_map} visualizes membership maps, computed by
\begin{equation}
\big[\operatorname{argmax}(\bfp_{1}), \operatorname{argmax}(\bfp_{2}), \dots, \operatorname{argmax}(\bfp_{HW})\big]
\end{equation}
where each element is the index of the region query that the corresponding pixel most likely belongs to. We see that each image is partitioned into $M$ meaningful regions. Thus, $F_s$ in \eqref{eq:regional} represents the regional membership of each pixel and is used to make the feature map $X$ more discriminative.

To produce the attention matrix $A$ reliably, we design a novel loss function in Section \ref{subsec:loss}. The default $M$ is 8.

\vspace*{0.1cm}
\noindent {\bf Compositional feature extraction:}
For each line combination, we generate three types of feature maps to extract a compositional feature map. First, we generate a binary mask $B \in \mathbb{R}^{HW}$ for each line combination. Each element in $B$ is 1 if the corresponding pixel belongs to any line in the combination, and 0 otherwise. Then, we decompose the combined feature into a line feature map $X_l$ and a region feature map $X_r$ by
\begin{equation}
    X_l = X \otimes B, \,\,\, X_r = X \otimes (1-B),
\end{equation}
where $\otimes$ is the element-wise multiplication. In other words, $X_l \in \mathbb{R}^{HW \times 2C}$ contains contextual information for the line pixels only, whereas $X_r \in \mathbb{R}^{HW \times 2C}$ does for the pixels strictly inside the regions divided by the lines.

\begin{figure}
\begin{center}
\includegraphics[width=1\linewidth]{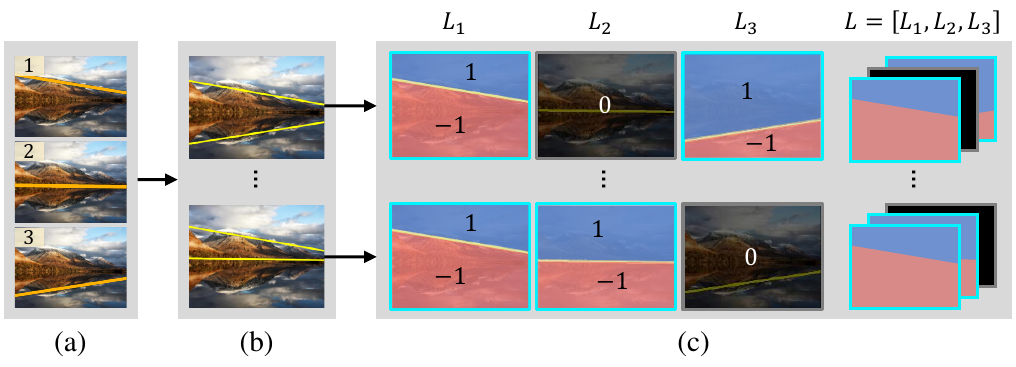}
\end{center}
\vspace{-0.5cm}
   \caption{Illustration of the line collection map generation when $K=3$: (a) reliable lines, (b) line combinations, and (c) line collection maps. }
\label{fig:line_collection}
\vspace{-0.3cm}
\end{figure}

Moreover, we generate a line collection map $L = [L_{1}, L_{2}, \dots, L_{K}] \in \mathbb{R}^{HW \times K}$, where $L_{k} \in \mathbb{R}^{HW}$ is a ternary mask for the $k$th reliable line, as illustrated in Figure~\ref{fig:line_collection}. If the $k$th reliable line does not belong to the line combination, all elements in $L_{k}$ are set to 0. Otherwise, each value in $L_k$ is set to either $1$ or $-1$ to indicate where the corresponding pixel is located between the two parts divided by the $k$th reliable line. In other words, $L_k$ informs how the $k$th line splits the image into two regions. We then produce a positional feature map $P \in \mathbb{R}^{HW \times 2C}$ by applying a series of fully connected layers to $L$. Then, we yield a compositional feature map $Z \in \mathbb{R}^{HW \times 6C}$ by
\begin{equation}
    Z = [X_l, X_r, P],
    \label{eq:comp_feature}
\end{equation}
which contains information about how the lines in the combination separate the image into multiple parts.

\vspace*{0.1cm}
\noindent {\bf Score regression:} Lastly, using a regressor, we estimate a composition score for each line combination. Then, we declare the line combination with the highest score as the optimal group of semantic lines. The regressor takes the compositional feature map $Z$ in (\ref{eq:comp_feature}) as input and predicts a composition score $s$ within $[0, 1]$. It is implemented using a bilinear interpolation layer and a series of 2D convolution layers and fully connected layers with the ReLU activation.

\subsection{Loss Functions}\label{subsec:loss}
To train SLCD, we design two loss functions.

\vspace*{0.1cm}
\noindent {\bf Semantic region separation loss:}
When two pixels $i$ and $j$ are located in distinct regions, they should be assigned to different region queries. In other words, their probability vectors $\bfp_{i}$ and $\bfp_{j}$ in the attention matrix $A$ should be far from each other. Let $X$ and $Y$ be the two regions divided by a ground-truth line. Their probability vectors are defined as
\begin{equation}
    \bfp_X = \frac{1}{|X|}\sum_{i \in X} \bfp_{i},  \quad\quad  \bfp_Y = \frac{1}{|Y|}\sum_{j \in Y} \bfp_{j},
    \label{eq:attention_vector}
\end{equation}
which should be also far from each other, for the ground-truth line divides the image into two semantic regions.

To measure the distance between probability vectors, we adopt the Kullback-Leibler divergence (KLD) \cite{cover2006elements}. Then, we define the semantic region segmentation (SRS) loss, by employing all ground-truth lines, as
\begin{equation}
    \mathcal{L}_{\rm SRS} = - \sum_{l=1}^{T} \big[D(\bfp_{X_l} \!\! \parallel \! \bfp_{Y_l}) + D(\bfp_{Y_l} \!\! \parallel \! \bfp_{X_l})\big]
    \label{eq:loss_KLD}
\end{equation}
where $D$ denotes the KLD,  $\bfp_{X_l}$ and $\bfp_{Y_l}$ are the regions divided by the $l$th ground-truth line, and $T$ is the number of ground-truth lines. Both $D(\bfp_{X_l}  \!\! \parallel \! \bfp_{Y_l})$ and $D(\bfp_{Y_l}  \!\! \parallel \! \bfp_{X_l})$ are computed because $D$ is not commutative. With $\mathcal{L}_{\rm SRS}$, it is possible to roughly segment an image into meaningful parts, as illustrated in Figure~\ref{fig:attn_map}, even though no ground-truth labels for semantic segmentation are used for training.

\vspace*{0.1cm}
\noindent {\bf Regression loss:} We also design a loss for regressing the composition score of each line combination. To this end, we employ the HIoU metric \cite{jin2021harmonious} that quantifies the structural layout of a line combination. Let $\bfc$ and $\bfc^{\star}$ denote a line combination and the ground-truth combination, respectively. We compute the HIoU between $\bfc$ and $\bfc^{\star}$ and use it as the ground-truth composition score ${\bar s}$ of $\bfc$. Then, the regression loss is defined as
\begin{equation}
\mathcal{L}_{\rm reg}=(s - {\bar s})^2
\end{equation}
where $s$ is the predicted score of $\bfc$. To reduce $\mathcal{L}_{\rm reg}$ effectively, a ranking loss $\mathcal{L}_{\rm rank}$ is also employed as in \cite{li2020composing}.

\begin{figure}
\begin{center}
\includegraphics[width=1\linewidth]{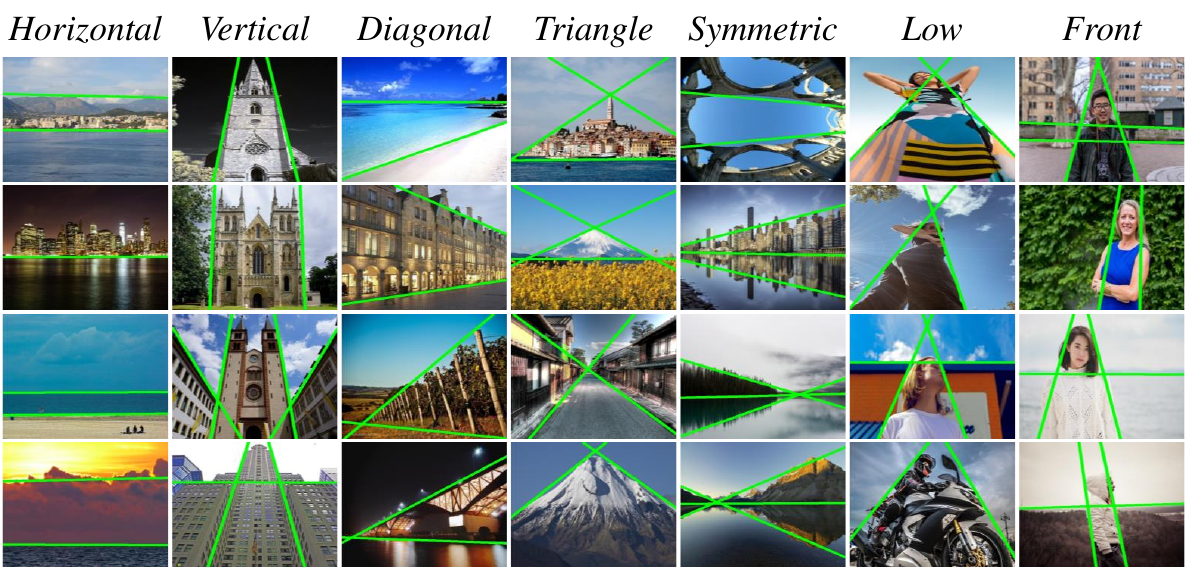}
\end{center}
\vspace{-0.5cm}
   \caption{The proposed CDL dataset contains 9,100 images in seven composition classes: \textit{Horizontal}, \textit{Vertical}, \textit{Diagonal}, \textit{Triangle}, \textit{Symmetric}, \textit{Low}, and \textit{Front}. The ground-truth semantic lines are depicted in green.}
   \vspace{-0.2cm}
\label{fig:CDL}
\end{figure}

\section{Experimental Results}
\subsection{Implementation Details}
We adopt ResNet50 \cite{he2016deep} as the encoder of the proposed SLCD. We use the AdamW optimizer \cite{loshchilov2017decoupled} with a learning rate $10^{-4}$, a weight decay of $10^{-4}$, $\gamma = 0.5$,  $\beta_1=0.9$, and $\beta_2=0.999$. We use a batch size of two for 400,000 iterations. Training images are resized to $480 \times 480$ and augmented by random horizontal flipping. We fix the number of line candidates, reliable lines, and region queries to $N=1024$, $K=8$, and $M=8$, respectively. We set $H=60$, $W=60$, and $C=96$.

\subsection{Datasets}

\noindent {\bf SEL \cite{lee2017semantic}:} It is the first dataset for semantic line detection. It contains 1,750 images, which are split into 1,575 training and 175 test images. Each semantic line is annotated by the coordinates of two endpoints on an image boundary. In SEL, most images are high-quality landscapes, in which semantic lines are often obvious.

\vspace*{0.1cm}
\noindent {\bf SEL{\_}Hard \cite{jin2020semantic}:} It contains 300 test images only, selected from the ADE20K segmentation dataset \cite{zhou2017scene}. It is challenging because of cluttered scenes or occluded objects.

\vspace*{0.1cm}
\noindent {\bf NKL \cite{zhao2021deep}:} It is a relatively large dataset of 5,200 training and 1,300 testing images. It includes both indoor and outdoor scenes.

\vspace*{0.1cm}
\noindent {\bf CDL:} We construct CDL to contain 7,100 scenes with diverse contents and compositions. It is split into 6,390 training and 710 test images. As in Figure~\ref{fig:CDL}, the images are categorized into seven composition classes: \textit{Horizontal}, \textit{Vertical}, \textit{Diagonal}, \textit{Triangle}, \textit{Symmetric}, \textit{Low}, and \textit{Front}. In \textit{Low} and \textit{Front}, humans and animals are essential parts of the image composition. In the other classes, most images are outdoor ones, as in the other datasets. To construct CDL, semantic lines were manually annotated in about 200 man-hours. More examples and the annotation process are provided in the appendix (Section B).

\begin{figure*}
\vspace*{-0.3cm}
\begin{center}
\includegraphics[width=1\linewidth]{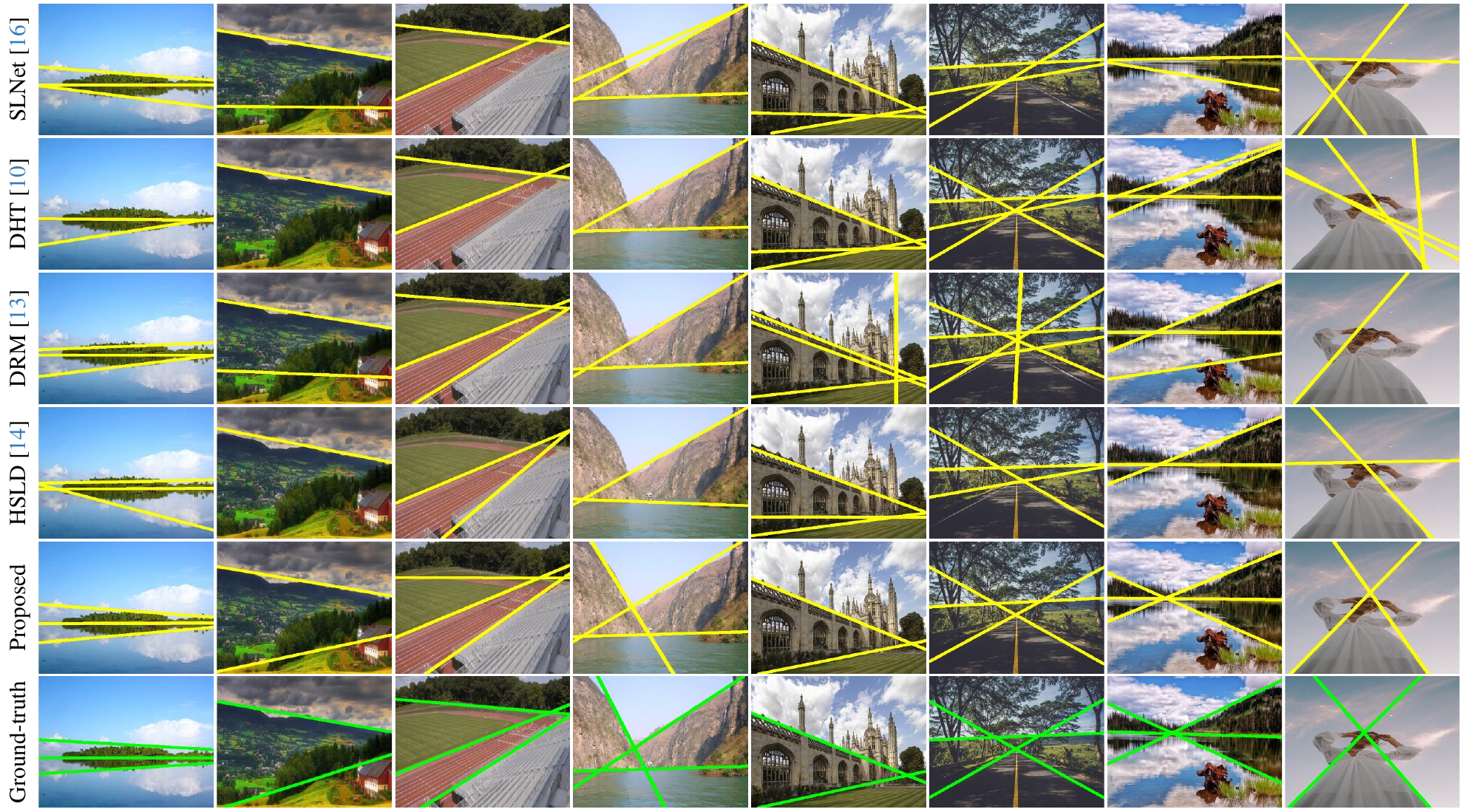}
\end{center}
\vspace{-0.5cm}
   \caption{Comparison of detected semantic lines. From the left, two images are selected from each of the SEL, SEL\_Hard, NKL, and CDL datasets.}
   \vspace{-0.2cm}
\label{fig:Comparison1}
\end{figure*}

\begin{table}[t]\centering
    \renewcommand{\arraystretch}{0.9}
    \caption
    {
        Comparison of the HIoU scores (\%) on the SEL, SEL\_Hard, NKL, and CDL datasets.
    }
    \vspace*{-0.15cm}
    \resizebox{1.0\linewidth}{!}{
    \begin{tabular}[t]{+L{2.2cm}^C{1.6cm}^C{1.6cm}^C{1.6cm}^C{1.6cm}}
    \toprule
    &  SEL  &  SEL\_Hard  &  NKL & CDL \\
    \midrule
        SLNet \cite{lee2017semantic}	& 77.87             & 59.71             & 65.49             & 57.78  \\
        DHT \cite{han2020eccv}		    & 79.62             & 63.39             & 69.08             & 63.24  \\
        DRM \cite{jin2020semantic}	    & 80.23             & \textbf{68.83}    & 67.42             & 63.96  \\
        HSLD \cite{jin2021harmonious}   & \underline{81.03} & 65.99             & \underline{74.29} & \underline{64.98}  \\
    \midrule
        Proposed                        & \textbf{84.09}    & \underline{68.15} & \textbf{76.21}    & \textbf{68.85}  \\
    \bottomrule
    \end{tabular}}
    \label{table:Comparison}
\end{table}

\subsection{Metrics}
As mentioned earlier, the harmony of detected lines is more important than individual lines in semantic line detection. Therefore, the main paper discusses only HIoU performances. The precision, recall, and f-measure results, which are metrics used in previous work, are compared in the supplemental document.

\vspace*{0.1cm}
\noindent {\bf HIoU:} An optimal group of semantic lines conveys a harmonious impression about the composition of an image. To assess the overall harmony of detection results, we use the HIoU metric \cite{jin2021harmonious}. It measures the consistency between the division of an image by detected lines and that by the ground truth. Let $S=\{s_1, s_2, \ldots, s_{N}\}$ and $T=\{t_1, t_2, \ldots, t_{M}\}$ be the regions divided by the detected lines and the ground-truth lines, respectively. Then, HIoU is computed as
\begin{equation}\label{eq:hiou}
    \textstyle
    {\rm HIoU} = \frac{\sum_{i=1}^{N} \max_{k}{{\rm IoU}(s_i, t_k)} + \sum_{j=1}^{M} \max_{k}{{\rm IoU}(t_j, s_k)}}{N+M}.
\end{equation}
Specifically, for each $s_i$, the most overlapping region $t_k$ is chosen and the intersection-over-union (IoU) is calculated. This is performed similarly for each $t_j$. Then, the HIoU score is given by the average of these bi-directionally matching IoUs.

\subsection{Comparative Assessment}
Table~\ref{table:Comparison} compares the HIoU scores of the proposed SLCD with those of the existing detectors \cite{lee2017semantic,han2020eccv,jin2020semantic,jin2021harmonious} on the SEL, SEL\_Hard, NKL, and CDL datasets. Also, Figure~\ref{fig:Comparison1} compares detection results qualitatively. The existing detectors miss correct lines or fail to remove redundant lines, yielding sub-optimal results. In contrast, SLCD detects semantic lines more precisely and represents the composition more reliably than the existing detectors do. More detection results are available in the appendix (Section D).

\vspace*{0.1cm}
\noindent {\bf Comparison on SEL:} In Table~\ref{table:Comparison}, SLCD outperforms the existing detectors on SEL. Compared with the second-best HSLD, SLCD yields a wide HIoU margin of 3.06. This indicates that SLCD finds an optimal group of semantic lines more effectively by processing all lines in each combination simultaneously, instead of performing pairwise comparisons in HSLD.

\vspace*{0.1cm}
\noindent {\bf Comparison on SEL\_Hard:} As done in \cite{jin2020semantic}, we conduct experiments on SEL\_Hard using the networks trained on the SEL dataset. In Table~\ref{table:Comparison}, SLCD ranks 2nd on SEL\_Hard. DRM provides a better result than SLCD, but it demands much higher complexity, as will be discussed in Section~\ref{subsec:analysis}.

\begin{table}[t]\centering
    \renewcommand{\arraystretch}{0.9}
    \caption
    {
        Comparison of the HIoU scores (\%) on the CDL dataset according to the number $M$ of region queries.
    }
    \vspace*{-0.15cm}
    \resizebox{0.95\linewidth}{!}{
    \begin{tabular}[t]{+L{1.3cm}^C{1.5cm}^C{1.5cm}^C{1.5cm}^C{1.5cm}}
    \toprule
        $M$     & 4 & 8 & 16 & 32\\
    \midrule
        HIoU    & 66.43 & 68.85 & 68.11 & 67.07 \\
    \bottomrule
    \end{tabular}}
    \vspace*{-0.2cm}
    \label{table:Ablation1}
\end{table}

\begin{figure}
\begin{center}
\includegraphics[width=1\linewidth]{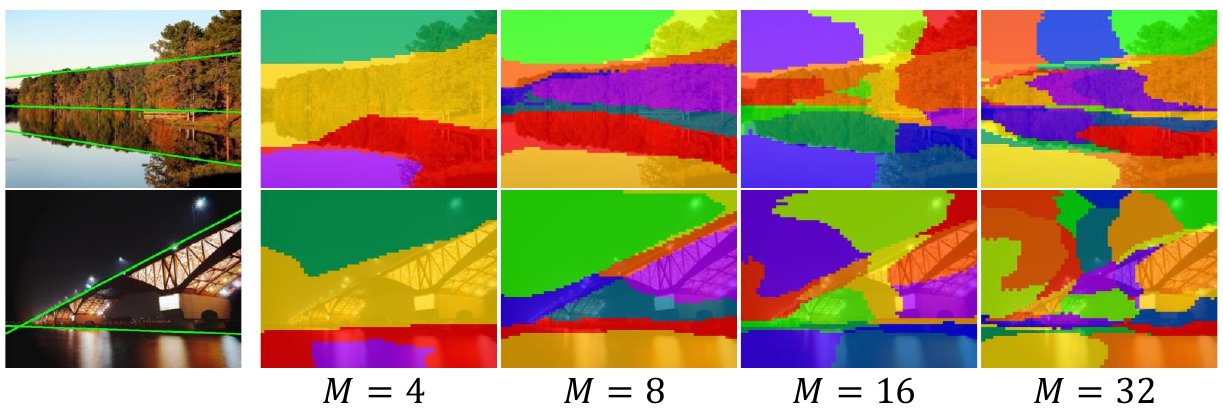}
\end{center}
\vspace{-0.5cm}
   \caption{Visualization of the membership maps according to $M$.}
   \vspace{-0.2cm}
\label{fig:Ablation_M}
\end{figure}

\vspace*{0.1cm}
\noindent {\bf Comparison on NKL:} SLCD surpasses all existing detectors. For example, its HIoU score is 1.92 points higher than the second-best HSLD.

\vspace*{0.1cm}
\noindent {\bf Comparison on CDL:} Table~\ref{table:Comparison} also lists HIoU scores on the proposed CDL dataset. For a fair comparison, we train the existing detectors on CDL using their publicly available source codes. We see that SLCD outperforms all existing detectors on CDL as well. SLNet, DHT, and DRM yield poor results because they do not consider the overall harmony of detected lines. HSLD is better than these detectors but is inferior to the proposed SLCD. We compare and discuss the detection results according to the composition classes in the appendix (Section D).

\subsection{Analysis}\label{subsec:analysis}
We conduct several ablation studies to analyze each component of the proposed SLCD.

\vspace*{0.1cm}
\noindent {\bf The number of $M$:}
Table \ref{table:Ablation1} lists the HIoU performances on the CDL dataset, according to the number $M$ of region queries. At the default $M=8$, SLCD achieves the best performance. When $M$ is smaller or bigger than 8, the performance degrades due to under- or over-segmentation of semantic parts, respectively, as shown in Figure~\ref{fig:Ablation_M}.

\vspace*{0.1cm}
\noindent {\bf Efficacy of $\mathcal{L}_{\rm SRS}$:}
In Table~\ref{table:Ablation2}, if the proposed SRS loss $\mathcal{L}_{\rm SRS}$ in \eqref{eq:loss_KLD} is excluded from the training, the performance drops by 2.14 points. It means that the composition analysis based on the SRS loss is essential for finding optimal line combinations.

\vspace*{0.1cm}
\noindent {\bf Efficacy of compositional feature extraction:}
SLCD extracts the compositional feature map $Z$ for each line combination, by processing the line feature map $X_l$, the region feature map $X_r$, and the positional feature map $P$ in \eqref{eq:comp_feature}. In Table~\ref{table:Ablation3}, Method \RomNum{1} uses the line feature map only, while \RomNum{2} uses the region feature map only. Method \RomNum{3} utilizes both feature maps. Method \RomNum{4}, which is the proposed SLCD, uses all three maps.

Method \RomNum{1} yields the worst result, for it uses only the contextual information near line pixels. Method \RomNum{2} is slightly better than Method \RomNum{1}, by exploiting the regional information. Using both line and region feature maps, \RomNum{3} provides a better result. Moreover, \RomNum{4} further improves the performance significantly by utilizing the positional feature map. This is because the overall harmony is estimated more effectively, by combining the line structures in the positional feature map with scene contexts.

\begin{table}[t]\centering
    \renewcommand{\arraystretch}{0.9}
    \caption
    {
        Comparison of the HIoU scores (\%) on the CDL dataset according to the usage of the SRS loss $\mathcal{L}_{\rm SRS}$.
    }
    \vspace*{-0.15cm}
    \resizebox{0.95\linewidth}{!}{
    \begin{tabular}[t]{+L{1.6cm}^C{3.4cm}^C{3.0cm}}
    \toprule
    & w/o $\mathcal{L}_{\rm SRS}$ & Proposed \\
    \midrule
        HIoU & 66.71 & 68.85  \\
    \bottomrule
    \end{tabular}}
    \label{table:Ablation2}
\end{table}

\begin{table}[t]\centering
    \renewcommand{\arraystretch}{0.95}
    \caption
    {
        Ablation studies for the compositional feature extraction on the CDL dataset.
    }
    \vspace*{-0.2cm}
    \resizebox{0.95\linewidth}{!}{
    \begin{tabular}[t]{+C{0.2cm}^L{0.4cm}^C{1.4cm}^C{1.4cm}^C{1.4cm}^C{2.0cm}}
    \toprule
                    & & Line       & Region     & Positional & HIoU\\
    \midrule
        \RomNum{1}  & & \checkmark &  		    &            & 65.12  \\
        \RomNum{2}  & & 		   & \checkmark &			 & 65.74  \\
        \RomNum{3}  & & \checkmark & \checkmark & 		     & 66.37  \\
        \RomNum{4}  & & \checkmark & \checkmark & \checkmark & 68.85  \\
    \bottomrule
    \end{tabular}}
    \label{table:Ablation3}
\end{table}

\vspace*{0.1cm}
\noindent {\bf Runtime:}
Table~\ref{table:runtime} compares the runtimes of SLCD and the existing detectors in seconds per frame (spf). We use a PC with AMD Ryzen 9 3900X CPU and NVIDIA RTX 2080 GPU. The proposed SLCD takes 0.114spf, adding up 0.030spf and 0.074spf for generating and assessing line combinations, respectively. DHT is the fastest detector, but it is inferior to SLCD on all datasets. On the other hand, even though DRM performs better on SEL\_Hard, it is about 8.4 times slower than SLCD.

\section{Applications}

We apply the proposed SLCD to three vision tasks: dominant vanishing point detection, reflection symmetry axis detection, and composition-based image retrieval. Note that the first two tasks were considered in \cite{jin2020semantic}. Due to the page limit, more details and results are described in the appendix (Section E).

\subsection{Dominant Vanishing Point Detection}\label{subsec:VPD}
A vanishing point (VP) facilitates understanding of the 3D geometric structure. We apply the proposed SLCD  to detect dominant VPs, by identifying vanishing lines. We use the AVA landscape dataset \cite{zhou2017detecting}, in which two dominant parallel lines are annotated for each image. To detect a dominant VP, we generate line combinations containing two lines only. We then find the best combination, whose intersecting point is declared as a VP. Table \ref{table:VPD} compares this VP detection scheme with the existing line-based VP detectors in \cite{zhou2017detecting,jin2020semantic}. The angle accuracies (AAs) \cite{zhou2019neurvps} are compared. We see that SLCD is better than the existing detectors, except that its AA$1^\circ$ score is lower than that of Zhou \etal \cite{zhou2017detecting}. Figure~\ref{fig:VPD} shows some VP detection results.

\begin{table}[t]\centering
    \renewcommand{\arraystretch}{0.95}
    \caption
    {
        Runtime comparison of the proposed SLCD and the existing detectors. The processing times are reported in seconds per frame (spf).
    }
    \vspace*{-0.15cm}
    \resizebox{0.97\linewidth}{!}{
    \begin{tabular}[t]{+C{1.6cm}^C{1.6cm}^C{1.6cm}^C{1.6cm}^C{1.5cm}}
    \toprule
     SLNet \cite{lee2017semantic} & DHT \cite{han2020eccv} & DRM \cite{jin2020semantic} & HSLD \cite{jin2021harmonious} & Proposed \\
    \midrule
       0.136 & 0.033 & 0.952 & 0.046 & 0.114  \\
    \bottomrule
    \end{tabular}}
    \label{table:runtime}
\end{table}

\begin{table}[t]\centering
    \renewcommand{\arraystretch}{0.9}
    \caption
    {
        Comparison of the AA scores (\%) of VP detection.
    }
    \vspace*{-0.2cm}
    \resizebox{0.97\linewidth}{!}{
    \begin{tabular}[t]{+L{3.2cm}^C{1.6cm}^C{1.6cm}^C{1.6cm}}
    \toprule
                                                    & AA$1^{\circ}$     & AA$2^{\circ}$     & AA$10^{\circ}$  \\
    \midrule
        Zhou \etal \cite{zhou2017detecting}	        & \textbf{18.5}     & \underline{33.0}  & 60.0  \\
        Jin \etal \cite{jin2020semantic}            & 8.6               & 22.9              & \underline{68.3}  \\
    \midrule
        Proposed                                    & \underline{16.6}  & \textbf{36.9}     & \textbf{78.3}  \\
    \bottomrule
    \end{tabular}}
    \label{table:VPD}
\end{table}

\begin{figure}
\begin{center}
\includegraphics[width=1\linewidth]{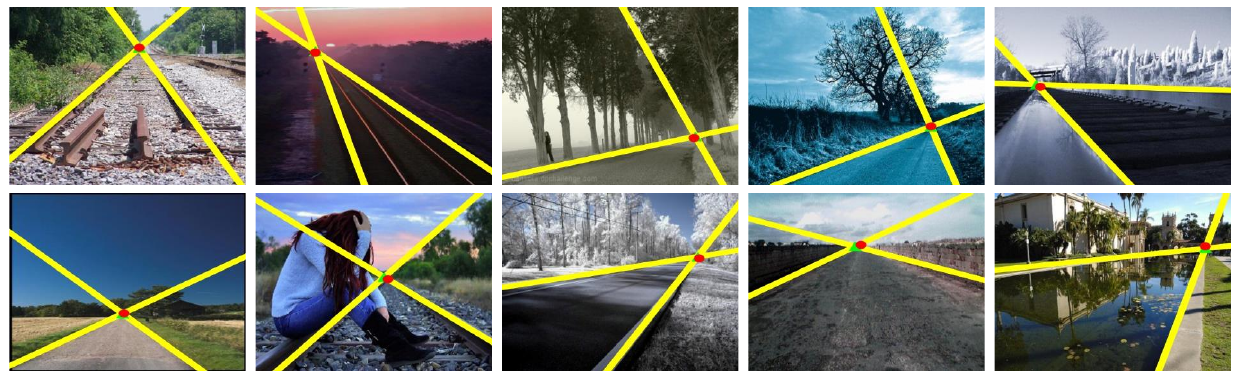}
\end{center}
\vspace{-0.4cm}
   \caption{Detected dominant parallel lines and their intersections are depicted by yellow lines and green triangles, respectively. The ground-truth VPs are depicted by red dots.}
   \vspace{-0.2cm}
\label{fig:VPD}
\vspace{-0.1cm}
\end{figure}

\subsection{Reflection Symmetry Axis Detection}\label{subsec:SYM}
\vspace{-0.15cm}
Reflection symmetry is a visual cue for analyzing object shapes and patterns. Its detection is challenging because the symmetry is only implicit in many cases. We test the proposed SLCD on three datasets: ICCV \cite{funk2017}, NYU \cite{cicconet2017finding}, and SYM\_Hard \cite{jin2020semantic}. In these datasets, each image contains a single reflection symmetry axis. Thus, in the proposed SLCD, each of the $K$ reliable lines is regarded as a line combination. Table \ref{table:SYM} compares the AUC\_A scores \cite{lee2017semantic} of the proposed SLCD and the conventional techniques \cite{cicconet2017_nyu,elawady2017,cicconet2017finding,loy2006detecting,jin2020semantic}. SLCD outperforms all these techniques on all datasets. Figure~\ref{fig:SYM} shows some detection results together with the membership maps. We see that regional parts are symmetrically divided along implied axes, enabling SLCD to identify those axes effectively.

\begin{table}[t]\centering
    \renewcommand{\arraystretch}{0.9}
    \caption
    {
        Comparison of the AUC\_A scores (\%) of symmetry axis detection.
    }
    \vspace*{-0.2cm}
    \resizebox{0.95\linewidth}{!}{
    \begin{tabular}[t]{+L{3.2cm}^C{1.6cm}^C{1.6cm}^C{1.6cm}}
    \toprule
                                                    & ICCV  & NYU   & SYM\_Hard \\
    \midrule
        Cicconet \etal \cite{cicconet2017_nyu}	    & 80.80 & 82.85 & 68.99  \\
        Elawady \etal \cite{elawady2017}		    & 87.24 & 83.83 & 73.90  \\
        Cicconet \etal \cite{cicconet2017finding}	& 87.38 & 87.64 & 81.04  \\
        Loy \& Eklundh \cite{loy2006detecting}      & 89.77 & 90.85 & 81.99  \\
        Jin \etal \cite{jin2020semantic}            & \underline{90.60} & \underline{92.78} & \underline{84.73}  \\
    \midrule
        Proposed                                    & \textbf{93.15} & \textbf{93.51} & \textbf{88.03}  \\
    \bottomrule
    \end{tabular}}
\label{table:SYM}
\vspace{-0.3cm}
\end{table}

\begin{figure}
\begin{center}
\includegraphics[width=1\linewidth]{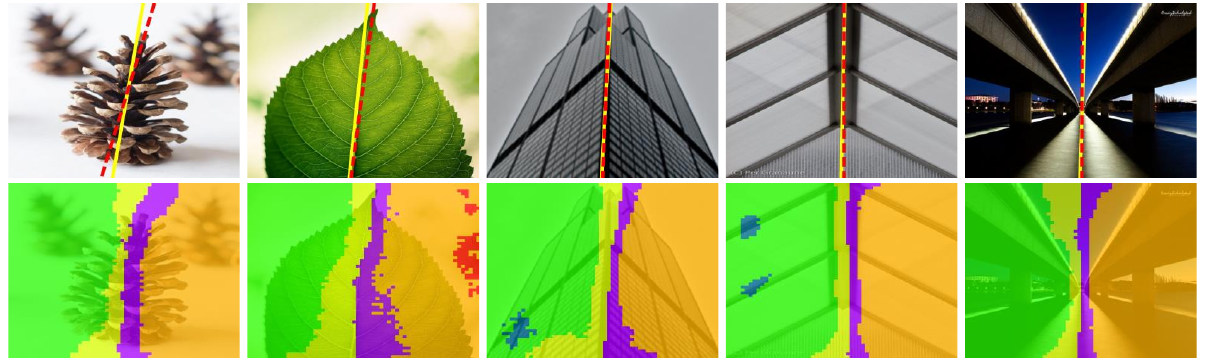}
\end{center}
\vspace{-0.6cm}
   \caption{Symmetry axis detection results. In the top row, the ground-truth and predicted axes are depicted by dashed red and solid yellow lines, respectively. The membership maps are also visualized in the bottom row.}
\label{fig:SYM}
\vspace{-0.4cm}
\end{figure}

\subsection{Composition-Based Image Retrieval}\label{subsec:RET}
\vspace{-0.15cm}
Existing image retrieval techniques focus on the visual contents of a query image to find similar images in a database \cite{noh2017large,radenovic2018fine,gordo2017end,smeulders2000content}. In this work, however, we attempt to discover images with similar compositions to a query image, \ie composition-based image retrieval.

We test the proposed SLCD on the Oxford 5k and Paris 6k retrieval dataset \cite{radenovic2018revisiting}. We first detect semantic lines in every image, while storing the positional feature map $P$ in (\ref{eq:comp_feature}). We filter out some images whose composition scores are lower than a threshold since a low score indicates a low-quality image with inharmonious composition in general. Then, for a randomly selected query image, we compute the $\ell_2$-distances between the positional feature maps $P$  of the query and the remaining images. We determine the images with the smallest distances as retrieval results.

\begin{figure}
\begin{center}
\includegraphics[width=1\linewidth]{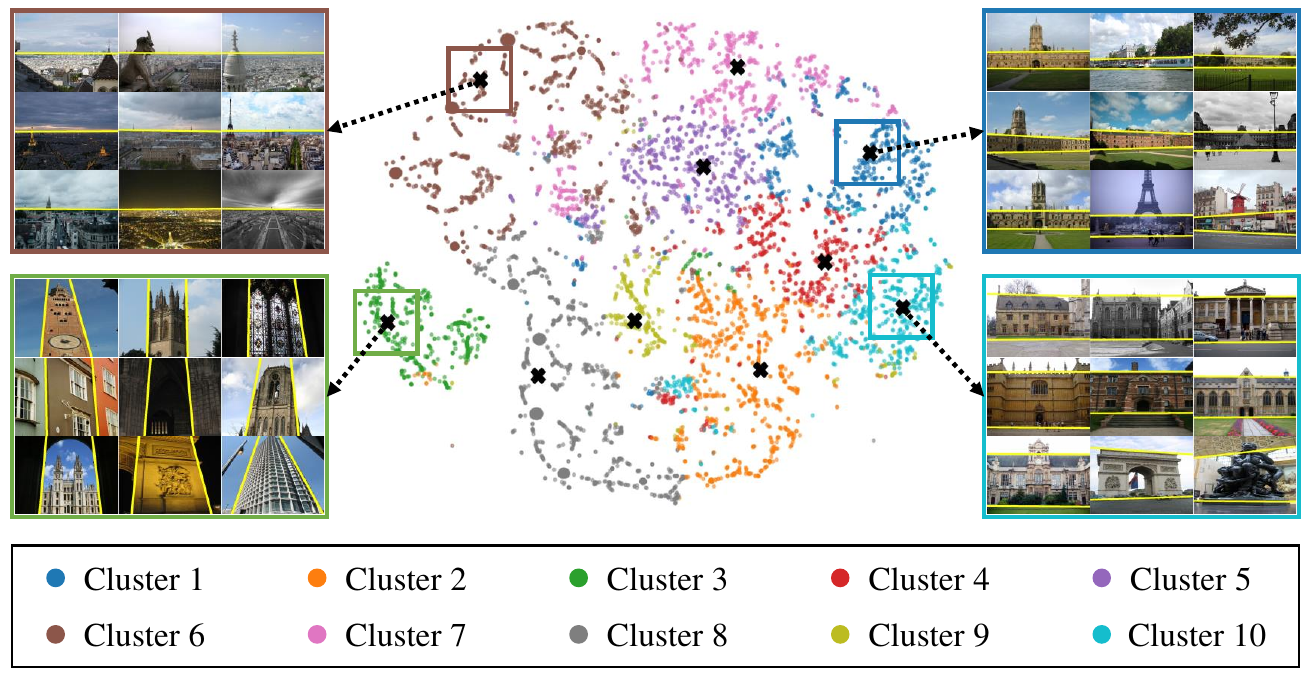}
\end{center}
\vspace{-0.6cm}
   \caption{t-SNE visualization \cite{van2008visualizing} of the feature space for the Oxford 5k and Paris 6k dataset \cite{radenovic2018revisiting}.}
\label{fig:tSNE}
\vspace{-0.4cm}
\end{figure}

Figure~\ref{fig:RTR} shows the top-4 retrieval results for four query images. We see that SLCD returns structurally similar images to the queries. This means that the positional feature maps $P$, containing the structural information of semantic lines, can be used to compute the compositional differences. Based on this observation, we perform data clustering by employing k-means \cite{hartigan1979algorithm} on the feature space of $P$. Figure~\ref{fig:tSNE} is  t-SNE visualization \cite{van2008visualizing} of the clustering results. The nine images nearest to each centroid are also shown. Note that images with similar composition are grouped into the same cluster, confirming the efficacy of SLCD in the compositional analysis of images.

\begin{figure}
\begin{center}
\includegraphics[width=1\linewidth]{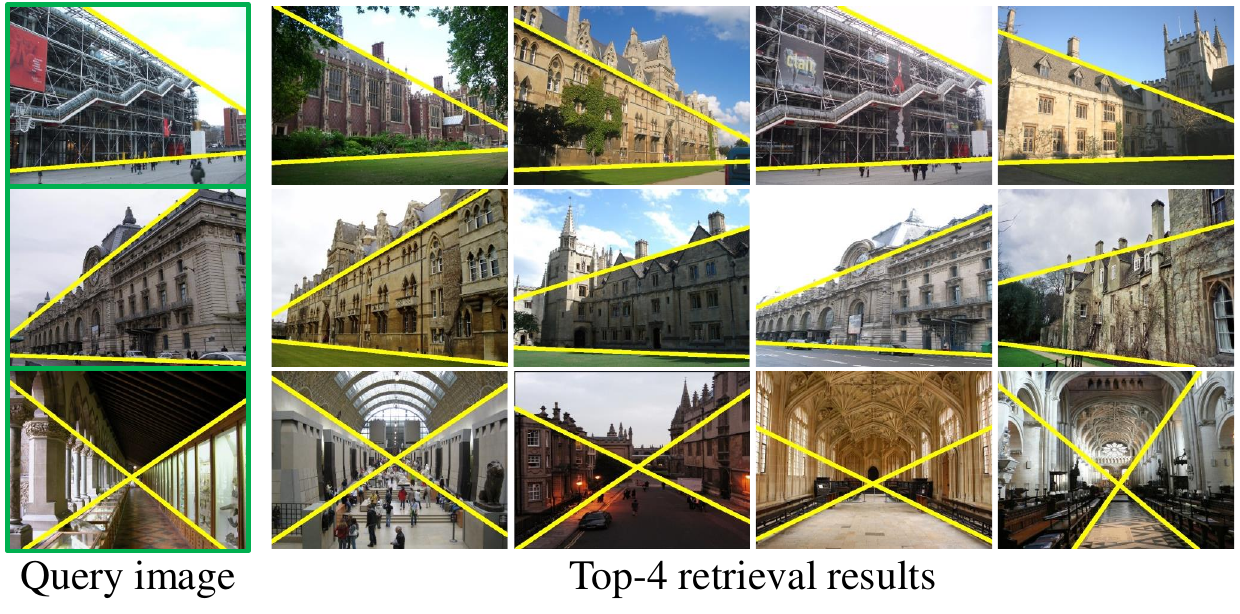}
\end{center}
\vspace{-0.7cm}
   \caption{Composition-based retrieval results for query images.}
\label{fig:RTR}
\vspace{-0.2cm}
\end{figure}

\vspace{-0.1cm}
\section{Conclusions}
\vspace{-0.15cm}
We proposed a novel semantic line detector, SLCD, which processes a combination of lines at once to estimate the overall harmony reliably. We first generated all possible line combinations from reliable lines. Then, we estimated the score of each line combination and determined the best combination. Experimental results demonstrated that the proposed SLCD can detect semantic lines reliably on existing datasets, as well as on the new dataset CDL. Furthermore, SLCD can be successfully used in vanishing point detection, symmetry axis detection, and image retrieval.

\vspace{-0.2cm}
\section*{Acknowledgements}
\vspace{-0.2cm}
This work was conducted by Center for Applied Research in Artificial Intelligence (CARAI) grant funded by DAPA and ADD (UD230017TD) and was supported by the National Research Foundation of Korea (NRF) grants funded by the Korea government (MSIT) (No.~NRF-2021R1A4A1031864 and No.~NRF-2022R1A2B5B03002310).

\clearpage

{\small
\bibliographystyle{ieeenat_fullname}
\bibliography{2024_CVPR_JWKO_arxiv}
}

\clearpage
\onecolumn
\begin{appendices}
\renewcommand{\thesection}{\Alph{section}}

\section{Implementation Details}\label{implement}
\subsection{Line detector}
\begin{figure}[h]
\begin{center}
\includegraphics[width=0.9\linewidth]{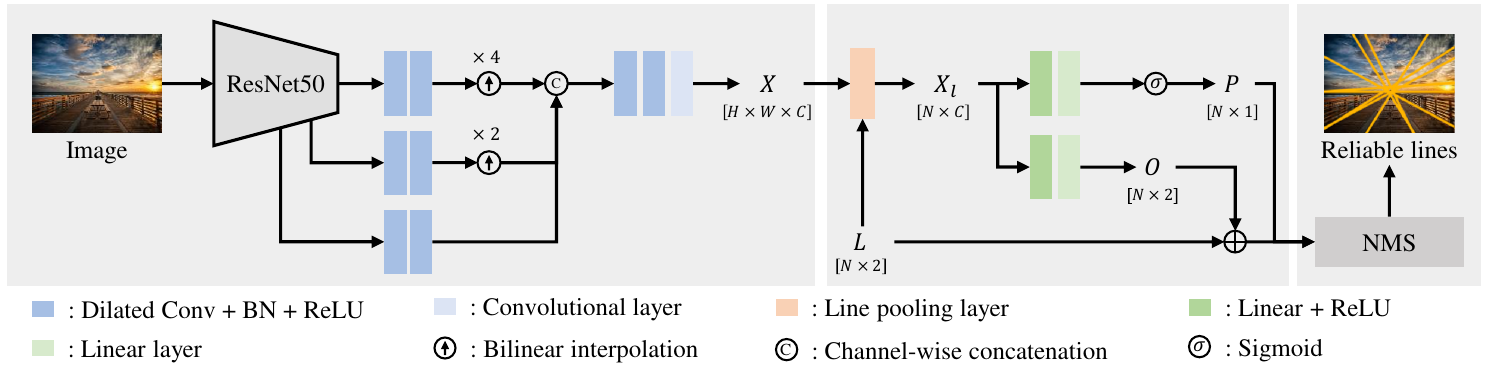}
\end{center}
  \vspace*{-0.4cm}
  \caption{The architecture of the modified line detector.}
  \label{fig:Detector}
\end{figure}

We design a line detector by modifying S-Net in \cite{jin2021harmonious} to select $K$ reliable lines from a set of line candidates. Figure~\ref{fig:Detector} shows the structure of the modified line detector performing three steps: encoding, classification/regression, and NMS.

\vspace*{0.1cm}
\noindent\textbf{Encoding:}
Given an image, the encoder extracts a convolutional feature map $X \in \mathbb{R}^{H \times W \times C}$. Specifically, we first extract multi-scale feature maps (the three coarsest feature maps) using ResNet50 \cite{he2016deep} as the backbone.
We then match each channel dimension to $C$ by applying a pair of $3 \times 3$ dilated convolutional layers with batch normalization and ReLU activation. We also match the resolutions of the two coarser maps to the finest one using bilinear interpolation and concatenate them. Finally, We obtain a combined feature map $X$ by applying two $3 \times 3$ dilated convolutional layers and one $3 \times 3$ convolutional layer.

\vspace*{0.2cm}
\noindent\textbf{Classification and regression:}
A line candidate, which is an end-to-end straight line in an image, can be parameterized by polar coordinates in the Hough space \cite{han2020eccv}. Let $l=(\rho, \theta)$ denote a line, where $\rho$ is the distance from the center of the image and $\theta$ is its angle from the $x$-axis. Then, we generate an initial set of $N$ line candidates, $L = [l_1, l_2, \dots, l_N] \in \mathbb{R}^{N\times 2}$, by quantizing $\rho$ and $\theta$ uniformly.
For the $N$ line candidates, we predict a probability vector $P \in \mathbb{R}^{N \times 1}$ and an offset matrix $O \in \mathbb{R}^{N \times 2}$. To this end, we extract a line feature matrix $X_{l} \in \mathbb{R}^{N \times C}$ of $L$ by employing a line pooling layer \cite{lee2017semantic}. Then, we feed $X_{l}$ into classification and regression layers to predict $P$ and $O$, respectively. More specifically, $P = \sigma(f_{1}(G))$ and $O = f_{2}(G)$, where $f_{1}$ and $f_{2}$ consist of two fully-connected layers, respectively, and $\sigma(\cdot)$ is the sigmoid function. Then, we update the line candidates by $\hat{L} = L + O$.

\vspace*{0.2cm}
\noindent\textbf{NMS:}
We select $K$ reliable lines from the updated set $\hat{L}$ based on the line probability matrix $P$. Specifically, we select the line with the highest probability and then remove the overlapping lines, whose $L_2$-distances in polar coordinates from the selected one are lower than a threshold. We iterate this process $K$ times to determine $K$ reliable lines.

\vspace*{0.2cm}
\noindent\textbf{Generating line combinations:}
From the $K$ reliable lines, we generate all $2^{K}$ line combinations. This process is done in parallel. Specifically, we compose a combination matrix $\bfC = [\bfc_{1}, \bfc_{2}, \dots, \bfc_{2^{K}}] \in \mathbb{R}^{K \times 2^{K}}$. Each column $\bfc_{i} \in \mathbb{R}^{K}$ is a boolean vector, where the $k$th element $\bfc_{i}^{k}$ is set to 1 if the $k$th reliable line belongs to the $i$th line combination, and 0 otherwise.

\clearpage
\subsection{SLCD}
We develop the semantic line combination detector (SLCD) to find the best line combination. The structure of SLCD is shown in Figure \red{3} in the main paper. It performs four steps: encoding, semantic feature grouping, compositional feature extraction, and score regression. Here, we describe the semantic feature grouping and score regression steps in detail.

\begin{figure}[h]
  \centering
  \includegraphics[width=1\linewidth]{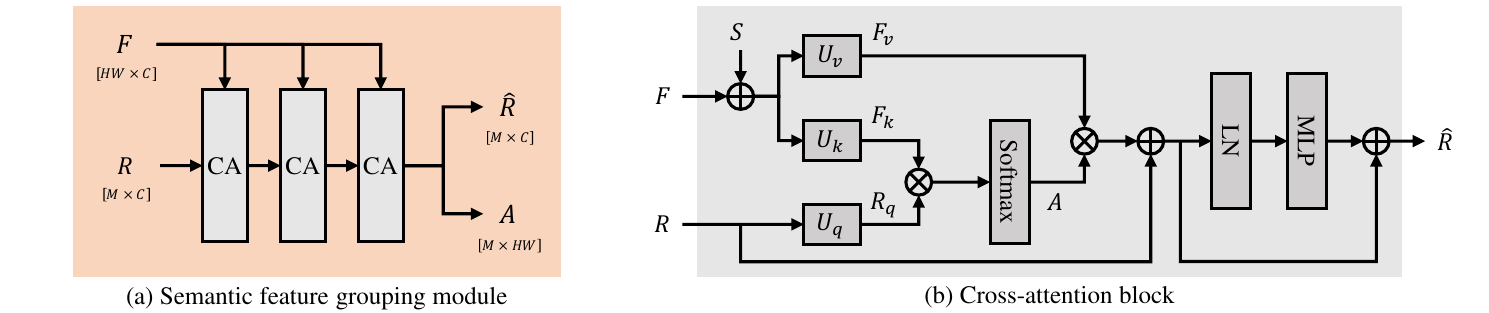}
  \caption{The architectures of the semantic feature grouping module and its cross-attention block.}
  \label{fig:SFG}
\end{figure}

\vspace*{0.1cm}
\noindent\textbf{Semantic feature grouping:}
Figure~\ref{fig:SFG} (a) and (b) show the semantic feature grouping module and its cross-attention (CA) block, respectively. In Figure~\ref{fig:SFG} (a), the module updates the region query matrix $R$ three times using CA blocks. Each CA block takes $F$ and $R$ as input and yields updated region query matrix $\hat{R}$. In the last block, it additionally outputs the attention matrix $A$. In Figure~\ref{fig:SFG} (b), LN and MLP denote layer normalization and multi-layer perceptron, respectively. The MLP consists of two fully connected layers with ReLU activation. Also, the projection matrices $U_{q}, U_{k}, U_{v} \in \mathbb{R}^{C \times C}$ are implemented by a fully connected layer.

\vspace*{0.1cm}
\noindent\textbf{Score regression:}
We assess each line combination using a regressor. The regressor takes the compositional feature map $Z \in \mathbb{R}^{HW \times 6C}$ as input and predicts a composition score $s$. More specifically, we first obtain a reduced feature map by $Z'=f(\psi(Z)) \in \mathbb{R}^{\frac{H}{4}\frac{W}{4} \times C}$, where $\psi$ is the bilinear interpolation to reduce the resolution by a factor of $\frac{1}{4}$. Also, $f$ consists of 2D convolution layers to reduce the channel dimension to $C$. We then estimate the score by $s=g(\bar{Z'})$, where $\bar{Z'}$ is a flattened $Z'$, and $g$ consists of a series of fully connected layers with ReLU activation.

\vspace*{0.5cm}
\subsection{Training and hyper-parameters}
Let us describe the training process of the modified line detector in detail.

\vspace*{0.1cm}
\noindent\textbf{Line detector:}
We generate a ground-truth (GT) line probability vector $\bar{P} \in \mathbb{R}^{N \times 1}$ and a GT offset matrix $\bar{O} \in \mathbb{R}^{N\times 2}$. Specifically, we set $\bar{P}_{l} = 1$ if the $L_2$-distance in polar coordinates between a line candidate and the best matching GT line is lower than a threshold, and $\bar{P}_l = 0$ otherwise. Also, we obtain $\bar{O}_{l}$ by computing the offset vector between a line candidate $l$ and its matching GT line. To train the modified line detector, we minimize the loss
\begin{equation}
    \mathcal{L} = \lambda_{1}\ell_{\rm cls}(P, \bar{P}) + \lambda_{2}\ell_{\rm reg}(O, \bar{O})
    \label{eq:loss_LD}
\end{equation}
where $\ell_{\rm cls}$ is the cross-entropy loss, and $\ell_{\rm reg}$ is the smooth $L_{1}$ loss. We set $\lambda_{1}$ and $\lambda_{2}$ to 1 and 5, respectively. We use the AdamW optimizer \cite{loshchilov2017decoupled} with an initial learning rate of $10^{-4}$ and halve it after every 80,000 iterations five times. We use a batch size of eight. Also, we resize training images to $480 \times 480$ and augment them via random horizontal flipping and random rotation between $-5^\circ$ and $5^\circ$.

\vspace*{0.1cm}
\noindent\textbf{Hyper-parameters:}
Table \ref{table:hyperparam} lists the hyper-parameters in the proposed algorithm. We will make the source codes publicly available.

\begin{table*}[h]\centering
    \renewcommand{\arraystretch}{0.9}
    \caption
    {
        Hyper-parameter settings.
    }
    \vspace*{-0.15cm}
    \footnotesize
    \begin{tabular}[t]{+L{1.6cm}^L{4.4cm}+L{1.6cm}^L{4.4cm}}
    \toprule
    Line detector & Description & SLCD & Description \\
    \midrule
    $H=60$ & Feature height & $H=60$ & Feature height \\
    $W=60$ & Feature width & $W=60$ & Feature width \\
    $C=96$ & \# of feature channels & $C=96$ & \# of feature channels \\
    $N=1024$ & \# of line candidates & $M=8$ & \# of region queries \\
    $K=8$ & \# of reliable lines & $\tau=1$ & scaling factor \\
    \bottomrule
    \end{tabular}
    \label{table:hyperparam}
    \vspace*{-0.1cm}
\end{table*}

\section{CDL Dataset}
\subsection{Data preparation}
Semantic lines represent the layout and composition of an image. Despite the close relationship between semantic lines and photographic composition, existing semantic line datasets do not consider image composition. To assess semantic line detection results for various types of composition, we construct a compositionally diverse semantic line dataset, called CDL, in which the photos are categorized into seven composition classes: \textit{Horizontal}, \textit{Vertical}, \textit{Diagonal}, \textit{Triangle}, \textit{Symmetric}, \textit{Low}, and \textit{Front}. In \textit{Low} and \textit{Front}, humans and animals are essential parts of the image composition. In the other classes, most images are outdoor ones, as in the other datasets. CDL contains 7,100 scenes split into 6,390 training and 710 test images. Figure~\ref{fig:CDL} shows example images and annotations in the proposed CDL.

\begin{figure}[h]
  \centering
  \includegraphics[width=1\linewidth]{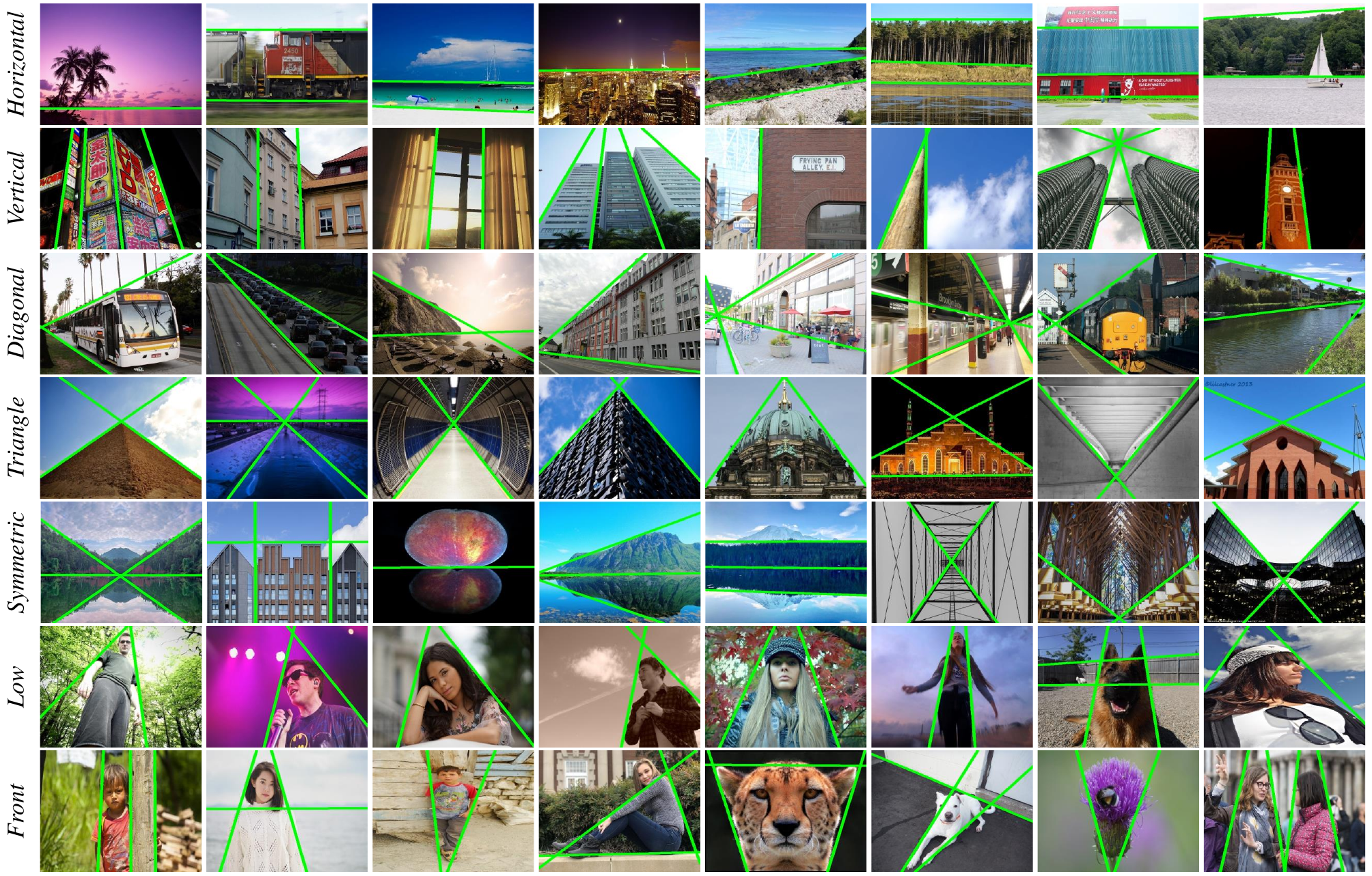}
  \caption{Example images and annotations in the proposed CDL dataset.}
  \label{fig:CDL}
\end{figure}

\clearpage
\subsection{Manual annotation}
\begin{figure}[h]
  \centering
  \includegraphics[width=1\linewidth]{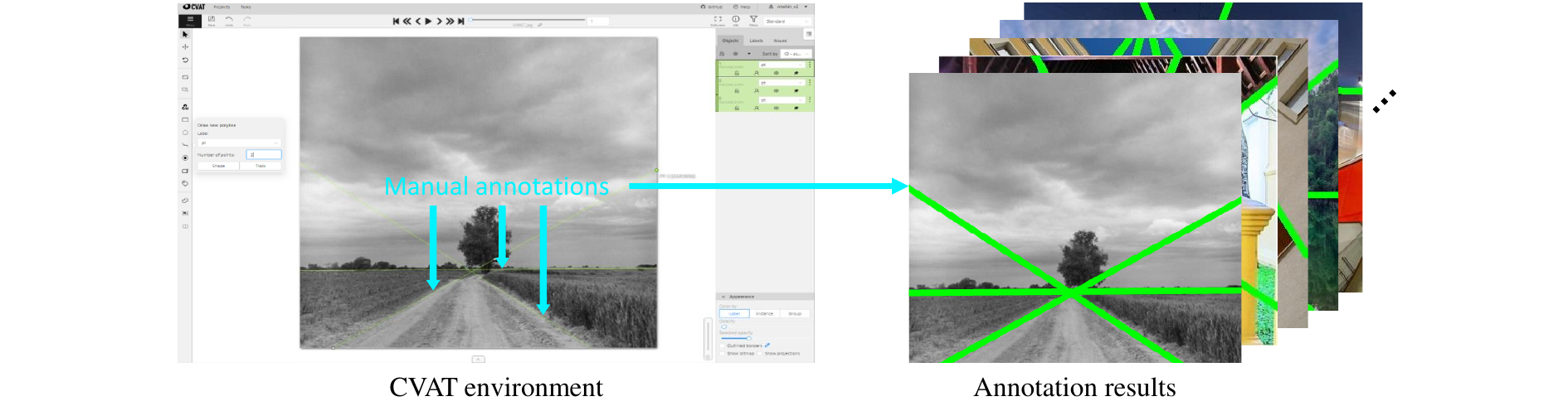}
  \vspace*{-0.6cm}
  \caption{Manual annotation process of semantic lines using CVAT. Ground-truth lines are depicted in green.}
  \label{fig:CVAT}
\end{figure}

For each image, we annotated the start and end points of each semantic line. We employed an annotation tool, CVAT \cite{cvat}, as shown in Figure~\ref{fig:CVAT}. We adjusted the coordinates of each line so that the lines represented the image composition optimally and harmoniously. Eight people inspected each annotated image, and unsatisfactory annotations were identified and re-annotated. The inspection was repeated until a consensus was made. The manual annotation process took more than 200 man-hours.

\vspace*{-0.2cm}
\section{More analysis}
\subsection{Performances}
Table~\ref{table:PRF1} reports the performances on the SEL and SEL\_Hard datasets. We compare the area under curve (AUC) performances of the precision, recall, and F-measure curves, which are denoted by AUC\_P, AUC\_R, and AUC\_F, respectively \cite{lee2017semantic}. Also, Table~\ref{table:PRF2} compares the EA-scores \cite{zhao2021deep} on the NKL and proposed CDL datasets. Note that, unlike HIoU \cite{jin2021harmonious}, AUC performances and EA-scores do not consider the overall harmony of detected lines. They only consider the positional accuracy of each detected line. We see that the proposed SLCD provides better results on most datasets in terms of detecting harmonious lines while accurately identifying individual lines.

\begin{table*}[h]\centering
    \renewcommand{\arraystretch}{0.9}
    \caption
    {
        Comparison of the AUC and HIoU on the SEL and SEL\_Hard datasets.
    }
    \vspace{-0.2cm}
    \centering
    {\footnotesize
    \begin{tabular}{L{1.5cm}C{1.4cm}C{1.4cm}C{1.4cm}C{1.4cm}C{0.01cm}C{1.4cm}C{1.4cm}C{1.4cm}C{1.4cm}}
    \toprule
                                    & \multicolumn{4}{c}{SEL}            & & \multicolumn{4}{c}{SEL\_Hard}  \\
    \cmidrule(l){2-5} \cmidrule(l){7-10}
                                    & AUC\_P & AUC\_R & AUC\_F & HIoU    & & AUC\_P & AUC\_R & AUC\_F & HIoU \\
    \midrule
    SLNet \cite{lee2017semantic}    & 80.72  & \underline{84.22}  & 82.43  & 77.87   & & 74.22  & 70.68  & 72.41  & 59.71 \\
    DHT \cite{han2020eccv}          & 87.74  & 80.25  & 83.83  & 79.62   & & 83.55  & 67.98  & 75.09  & 63.39 \\
    DRM \cite{jin2020semantic}      & 85.44  & \textbf{86.87}  & 86.15  & 80.23   & & 87.19  & \textbf{77.69}  & \textbf{82.17}  & \textbf{68.83} \\
    HSLD \cite{jin2021harmonious}   & \underline{89.61}  & 83.93  & \underline{86.68}  & \underline{81.03}   & & \underline{87.60}  & \underline{72.56}  & \underline{79.38}  & 65.99 \\
    \midrule
    Proposed                        & \textbf{95.27}  & 80.03  & \textbf{86.99}  & \textbf{84.09}   & & \textbf{90.32}  & 70.14  & 78.96  & \underline{68.15} \\
    \bottomrule
\end{tabular}
}
\vspace*{-0.4cm}
\label{table:PRF1}
\end{table*}

\begin{table*}[h]\centering
    \renewcommand{\arraystretch}{0.9}
    \caption
    {
        Comparison of the EA-scores and HIoU on the NKL and CDL datasets.
    }
    \vspace{-0.2cm}
    \centering
    {\footnotesize
    \begin{tabular}{L{1.5cm}C{1.4cm}C{1.4cm}C{1.4cm}C{1.4cm}C{0.01cm}C{1.4cm}C{1.4cm}C{1.4cm}C{1.4cm}}
    \toprule
                                    & \multicolumn{4}{c}{NKL}                   & & \multicolumn{4}{c}{CDL}  \\
    \cmidrule(l){2-5} \cmidrule(l){7-10}
                                    & Precision & Recall & F-measure & HIoU     & & Precision & Recall & F-measure & HIoU \\
    \midrule
    SLNet \cite{lee2017semantic}    & 74.41     & 74.50  & 74.45     & 65.49    & & 63.99     & 74.08  & 68.61     & 57.78 \\
    DHT \cite{han2020eccv}          & 78.76     & \textbf{83.17}  & \underline{80.88}     & 69.08    & & 72.02     & \textbf{79.49}  & \underline{75.52}     & 63.24 \\
    DRM \cite{jin2020semantic}      & 74.27     & 79.28  & 76.67     & 67.42    & & \underline{72.80}     & 72.13  & 72.46     & 63.96 \\
    HSLD \cite{jin2021harmonious}   & \underline{79.60}     & \underline{81.14}  & 80.36     & \underline{74.29}    & & 71.74     & 75.86  & 73.73     & \underline{64.98} \\
    \midrule
    Proposed                        & \textbf{84.39}     & 80.60  & \textbf{82.44}     & \textbf{76.21}    & & \textbf{75.69}     & \underline{78.41}  & \textbf{77.02}     & \textbf{68.85} \\
    \bottomrule
\end{tabular}
}
\label{table:PRF2}
\end{table*}

Table~\ref{table:Categorical} shows the HIoU scores on the proposed CDL dataset according to the composition classes. We train the existing detectors on CDL using their publicly available source codes. We see that the proposed SLCD outperforms all the existing techniques in all classes without exception. Among the seven classes, \textit{Low} and \textit{Front} are challenging ones because the complex boundaries of objects (humans or animals) make it difficult to identify the implied semantic lines. Qualitative results are shown in Section~\ref{subsec:CDL}.

\begin{table*}[h]\centering
    \renewcommand{\arraystretch}{0.9}
    \caption
    {
        Comparison of the HIoU scores (\%) according to the composition classes in the CDL dataset.
    }
    \vspace*{-0.15cm}
    \centering
    {\footnotesize
    \begin{tabular}[t]{L{1.5cm}C{1.3cm}C{1.3cm}C{1.3cm}C{1.3cm}C{1.3cm}C{1.3cm}C{1.3cm}C{1.4cm}}
    \toprule
    & \textit{Horizontal} & \textit{Vertical} & \textit{Diagonal} & \textit{Triangle} & \textit{Symmetric} & \textit{Low} & \textit{Front} & Total \\
    \midrule
    SLNet \cite{lee2017semantic}	& 72.67 & 51.32 & 64.52 & 47.04 & 69.18 & 46.87 & 49.25 & 57.78 \\
    DHT \cite{han2020eccv}		    & 76.02 & 59.70 & 64.87 & \underline{59.45} & 65.61 & 47.82 & 60.43 & 63.24 \\
    DRM \cite{jin2020semantic}	    & 75.86 & \underline{61.19} & 67.86 & 51.69 & 69.09 & 49.68 & \underline{63.41} & 63.96 \\
    HSLD \cite{jin2021harmonious}   & \underline{76.43} & 60.36 & \underline{69.01} & 58.35 & \underline{69.66} & \underline{54.91} & 60.75 & \underline{64.98} \\
    \midrule
    Proposed                        & \textbf{77.55} & \textbf{66.15} & \textbf{72.87} & \textbf{67.89} & \textbf{72.03} & \textbf{57.13} & \textbf{64.84} & \textbf{68.85} \\
    \bottomrule
\end{tabular}
}
    \label{table:Categorical}
\end{table*}

\vspace*{-0.2cm}
\subsection{Ablations}
\noindent\textbf{$N$ and $K$:}
As described in Section \red{3.1} in the main paper, to generate a manageable number of line combinations, we filter out redundant line candidates and obtain $K$ reliable lines. Table~\ref{table:Ablation1} lists the performances according to the number $N$ of line candidates and the number $K$ of reliable lines on CDL: (a) Recall (EA-score) rate of the line detector according to $N$, and (b) HIoU of the proposed SLCD according to $K$. The processing times are also reported in seconds per frame (spf).

In Table~\ref{table:Ablation1} (a), as $N$ increases, the recall rate gets higher. However,  too large values of $N$ make the training of the line detector challenging, yielding a lower recall rate at $N=1444$, compared to $N=1024$. Hence, we set $N=1024$. In Table~\ref{table:Ablation1} (b), as $K$ increases, the HIoU score gets higher. At $K=10$, SLCD achieves the highest HIoU, but it becomes too slow. It takes 4 times longer at $K=10$ than at $K=8$, whereas their HIoU gap is 0.27 only. As a tradeoff between performance and speed, we set $K=8$.

\begin{table*}[h]
    \caption{Comparison of the performance according to the number $N$ of line candidates and the number $K$ of reliable lines on CDL.}
    \vspace*{-0.15cm}
    \begin{subtable}{0.5\textwidth}\centering
        {\footnotesize
        \begin{tabular}{L{1cm}C{0.8cm}C{0.8cm}C{0.8cm}C{0.8cm}}
        \toprule
        $K=8$                 & \multicolumn{4}{c}{Line detector}    \\
        \cmidrule(l){2-5}
        $N \vartriangleright$ & 400 & 900 & 1024 & 1444         \\
        \midrule
        Recall                & 88.61  & 90.16  & 91.77 & 91.23 \\
        spf                   & 0.027  & 0.028  & 0.030 & 0.032 \\
        \bottomrule
        \end{tabular}}
        \vspace{0.1cm}
        \caption{Recall rate of the line detector according to $N$}
    \end{subtable}%
    \begin{subtable}{0.5\textwidth}\centering
        {\footnotesize
        \begin{tabular}{L{1.5cm}C{0.8cm}C{0.8cm}C{0.8cm}}
        \toprule
        $N =1024$             &  \multicolumn{3}{c}{SLCD} \\
        \cmidrule(l){2-4}
        $K \vartriangleright$ & 6 & 8 & 10                \\
        \midrule
         HIoU                 & 67.96 & 68.85 & 69.12     \\
         spf                  & 0.033 & 0.074 & 0.316     \\
        \bottomrule
        \end{tabular}}
        \vspace{0.1cm}
        \caption{HIoU of SLCD according to $K$.}
    \end{subtable}
    \vspace*{-0.15cm}
\label{table:Ablation1}
\vspace*{-0.3cm}
\end{table*}

\vspace*{-0.2cm}
\subsection{Visualizations}

\begin{figure}[h]
  \centering
  \includegraphics[width=0.98\linewidth]{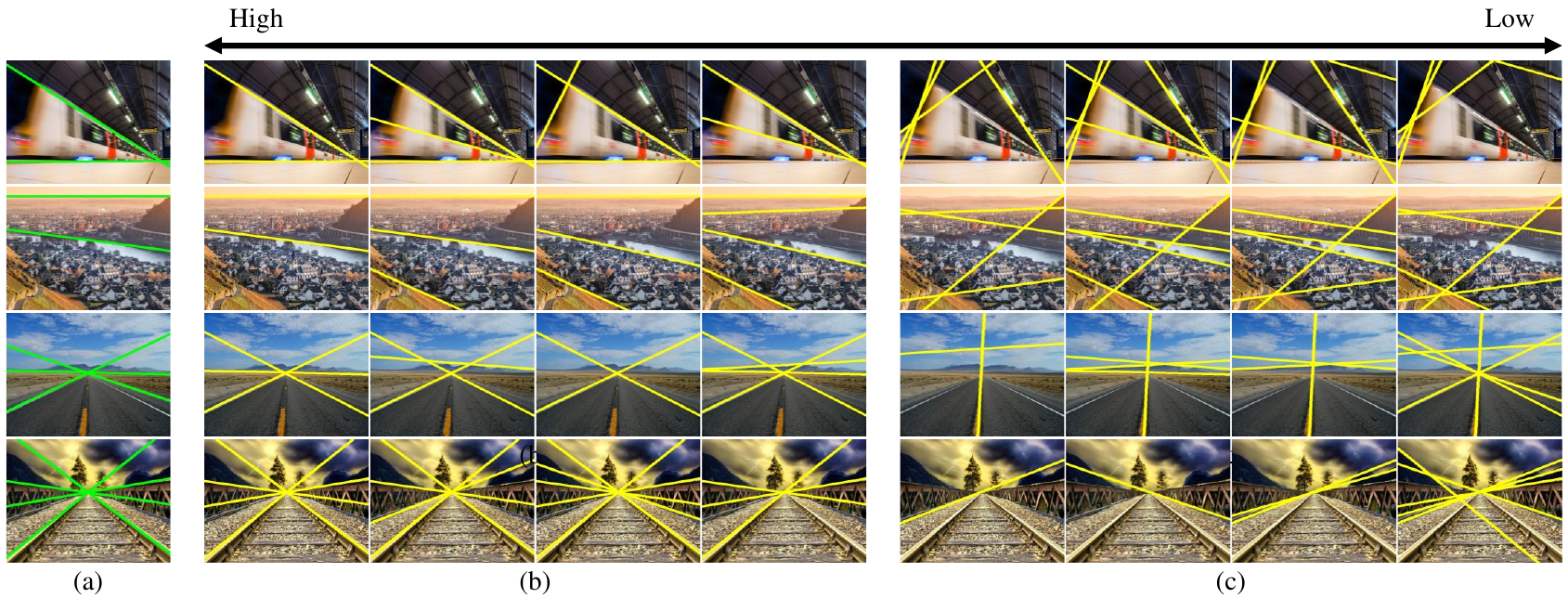}
  \vspace{-0.3cm}
  \caption{Sorting of line combinations based on predicted composition scores: (a) ground-truth lines, (b) top-4 line combinations, and (c) bottom-4 line combinations.}
  \label{fig:Sorting}
\vspace*{-0.5cm}
\end{figure}

\noindent\textbf{Sorting results according to composition scores:}
Note that we determine an optimal group of semantic lines in an image, by finding the line combination with the highest composition score, estimated by the proposed SLCD. Figure~\ref{fig:Sorting} (b) and (c) show examples of top-4 and bottom-4 line combinations, respectively. The top-4 detection results represent image composition faithfully. In contrast, the bottom-4 detection results represent it poorly. These examples indicate that the proposed SLCD evaluates each line combination reliably based on composition analysis.

\vspace*{-0.1cm}


\clearpage

\section{Qualitative Results}

\subsection{Comparison on SEL and SEL\_Hard}
\begin{figure}[h]
  \centering
  \includegraphics[width=0.98\linewidth]{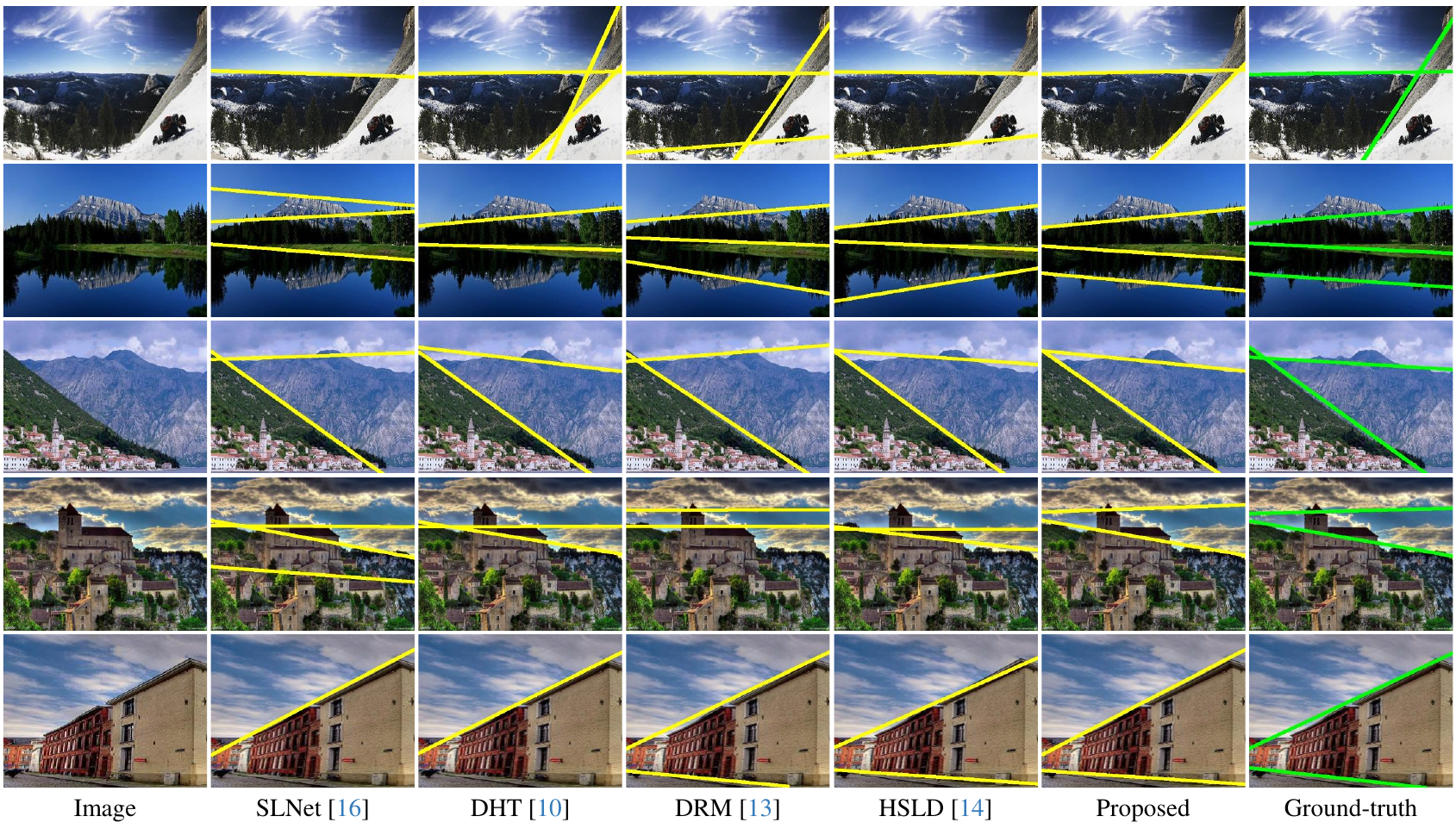}
  \vspace{-0.3cm}
  \caption{Comparison of semantic line detection results on the SEL dataset.}
  \label{fig:SEL}
\vspace{-0.6cm}
\end{figure}

\begin{figure}[h]
  \centering
  \includegraphics[width=0.98\linewidth]{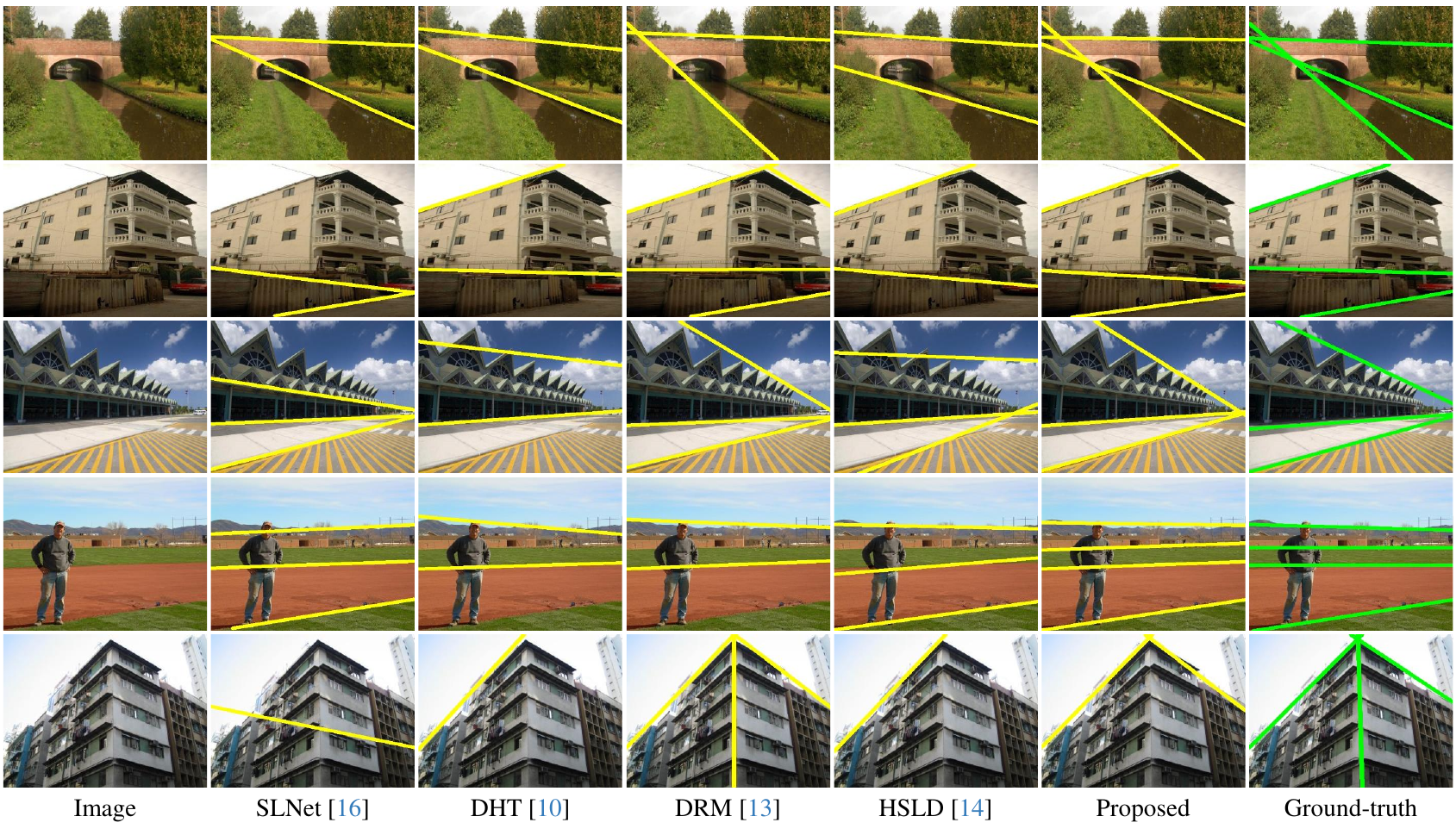}
  \vspace{-0.3cm}
  \caption{Comparison of semantic line detection results on the SEL\_Hard dataset.}
  \label{fig:SEL_Hard}
\vspace{-1cm}
\end{figure}

Figure~\ref{fig:SEL} and Figure~\ref{fig:SEL_Hard} compare the proposed SLCD with conventional algorithms on SEL and SEL\_Hard, respectively. SLCD provides better detection results than the conventional algorithms.

\clearpage

\subsection{Comparison on NKL}
\begin{figure}[h]
  \centering
  \includegraphics[width=0.98\linewidth]{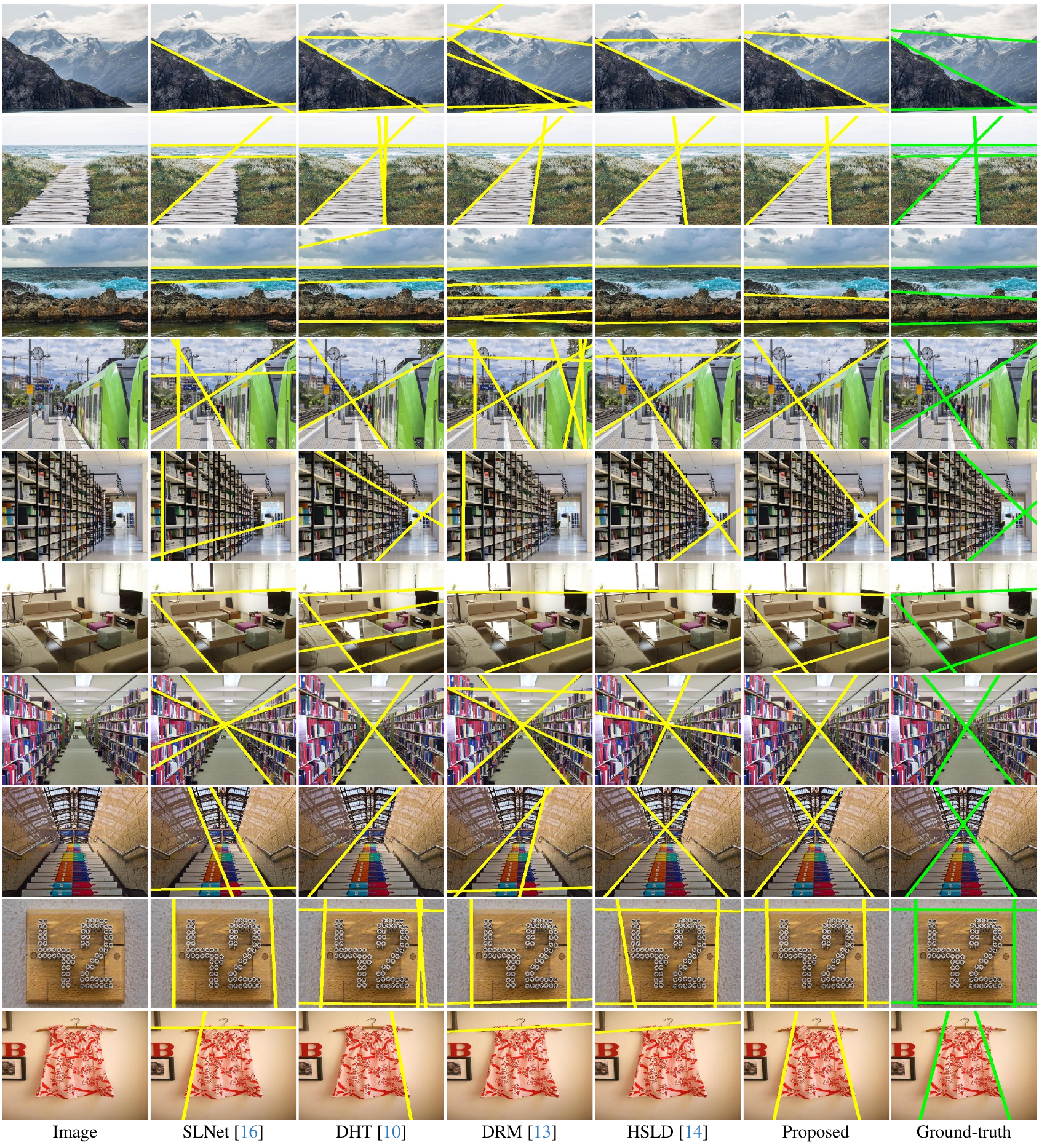}
  \caption{Comparison of semantic line detection results on the NKL dataset.}
  \label{fig:NKL}
\end{figure}

Figure~\ref{fig:NKL} compares the proposed SLCD with existing line detectors on NKL.

\clearpage

\subsection{Comparison on CDL}\label{subsec:CDL}
\begin{figure}[h]
  \centering
  \includegraphics[width=0.98\linewidth]{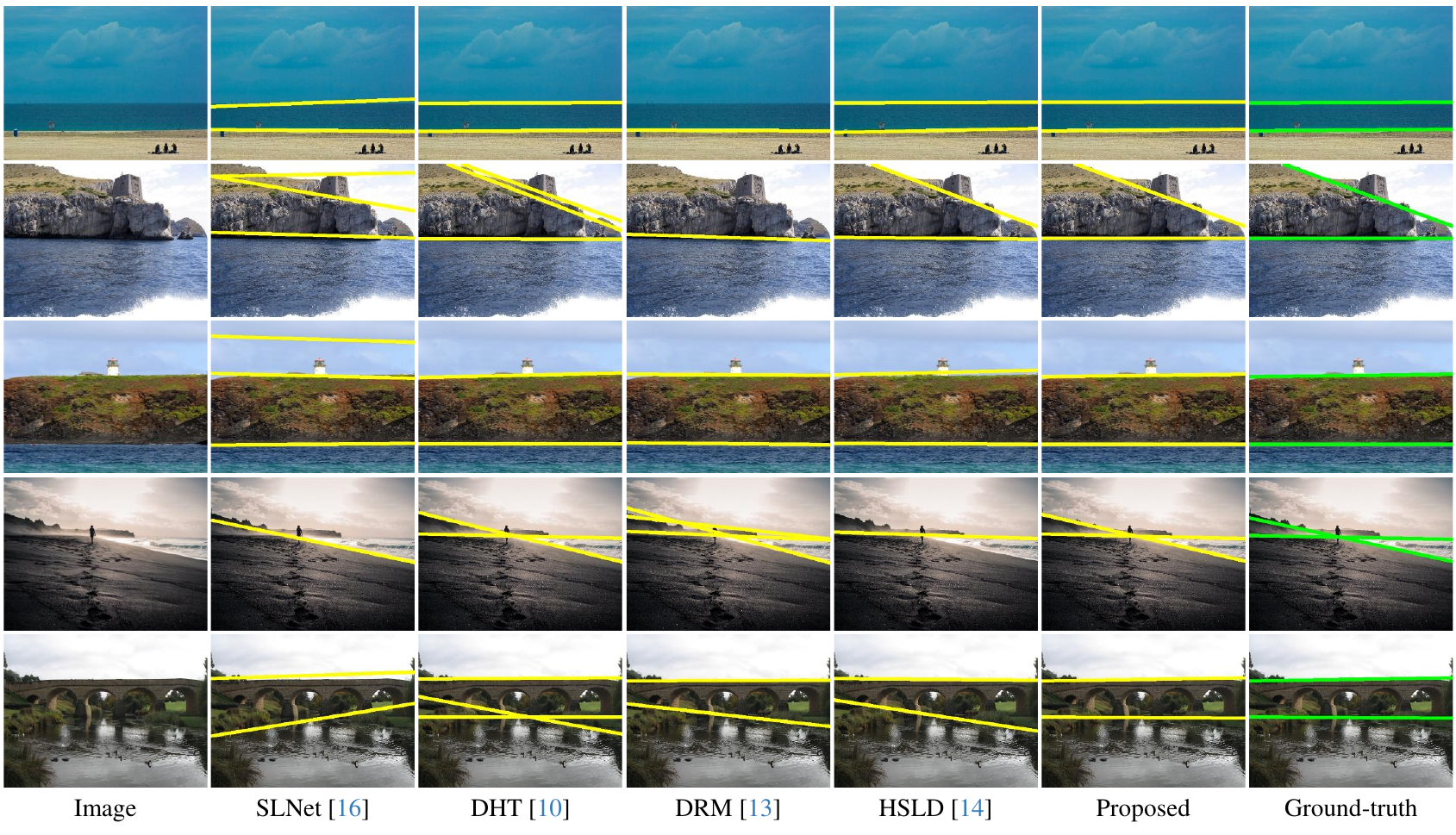}
  \vspace{-0.3cm}
  \caption{Comparison of semantic line detection results on the \textit{Horizontal} class in the CDL dataset.}
  \label{fig:CDL_Horizontal}
\vspace{-0.4cm}
\end{figure}

\begin{figure}[h]
  \centering
  \includegraphics[width=0.98\linewidth]{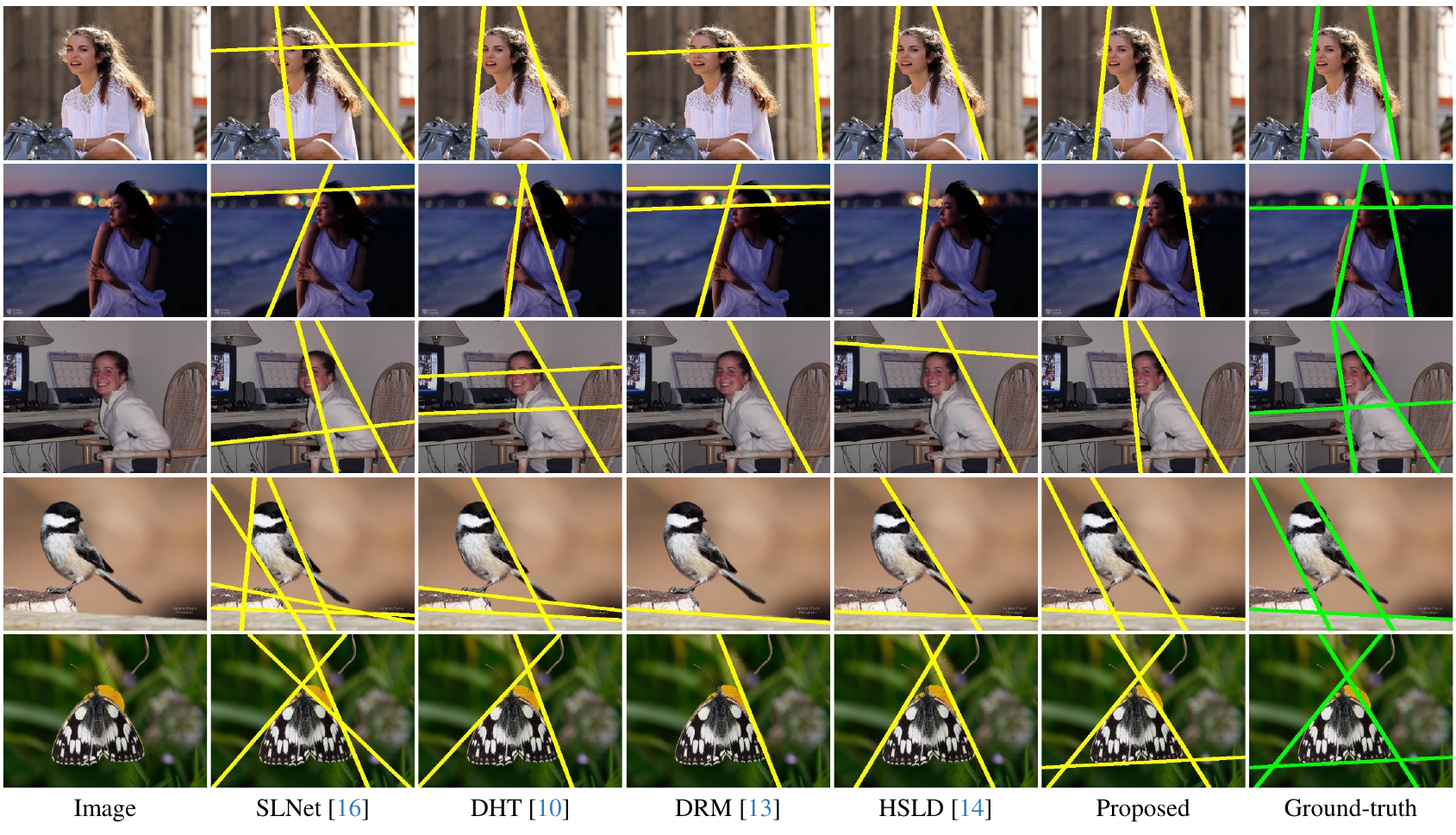}
  \vspace{-0.3cm}
  \caption{Comparison of semantic line detection results on the \textit{Front} class in the CDL dataset.}
  \label{fig:CDL_Front}
\end{figure}

Figures~\ref{fig:CDL_Horizontal}\red{$\sim$}\ref{fig:CDL_Front} compare semantic line detection results of the proposed CDL with those of the conventional algorithms on the \textit{Horizontal}, \textit{Vertical}, \textit{Diagonal}, \textit{Triangle}, \textit{Symmetric}, \textit{Low}, and \textit{Front} classes in the CDL dataset, respectively. We see that the proposed algorithm provides more reliable detection results consistently.

\begin{figure}[h]
  \centering
  \includegraphics[width=0.98\linewidth]{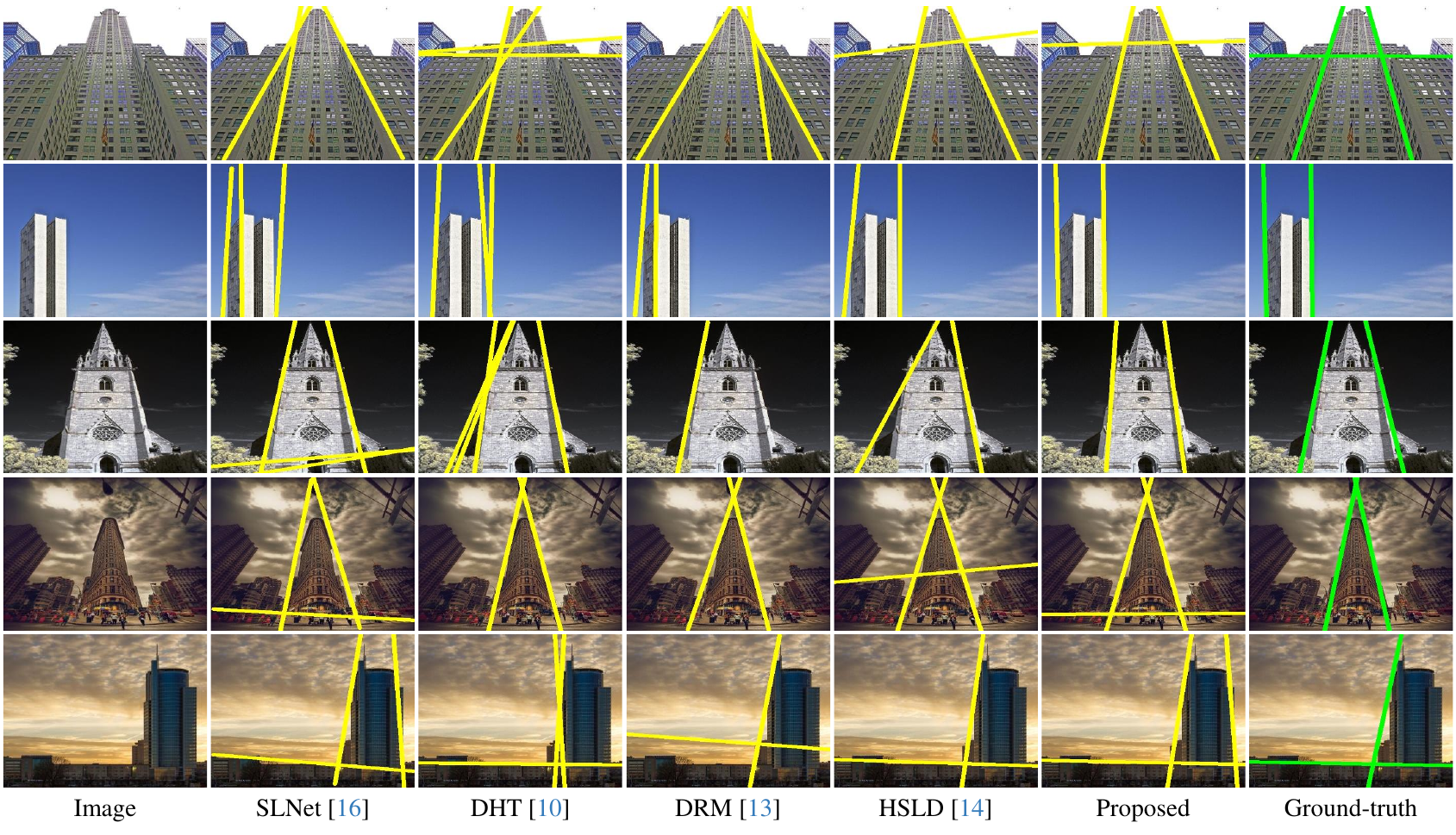}
  \vspace{-0.3cm}
  \caption{Comparison of semantic line detection results on the \textit{Vertical} class in the CDL dataset.}
  \label{fig:CDL_Vertical}
\vspace{-1cm}
\end{figure}

\clearpage

\begin{figure}[h]
  \centering
  \includegraphics[width=0.98\linewidth]{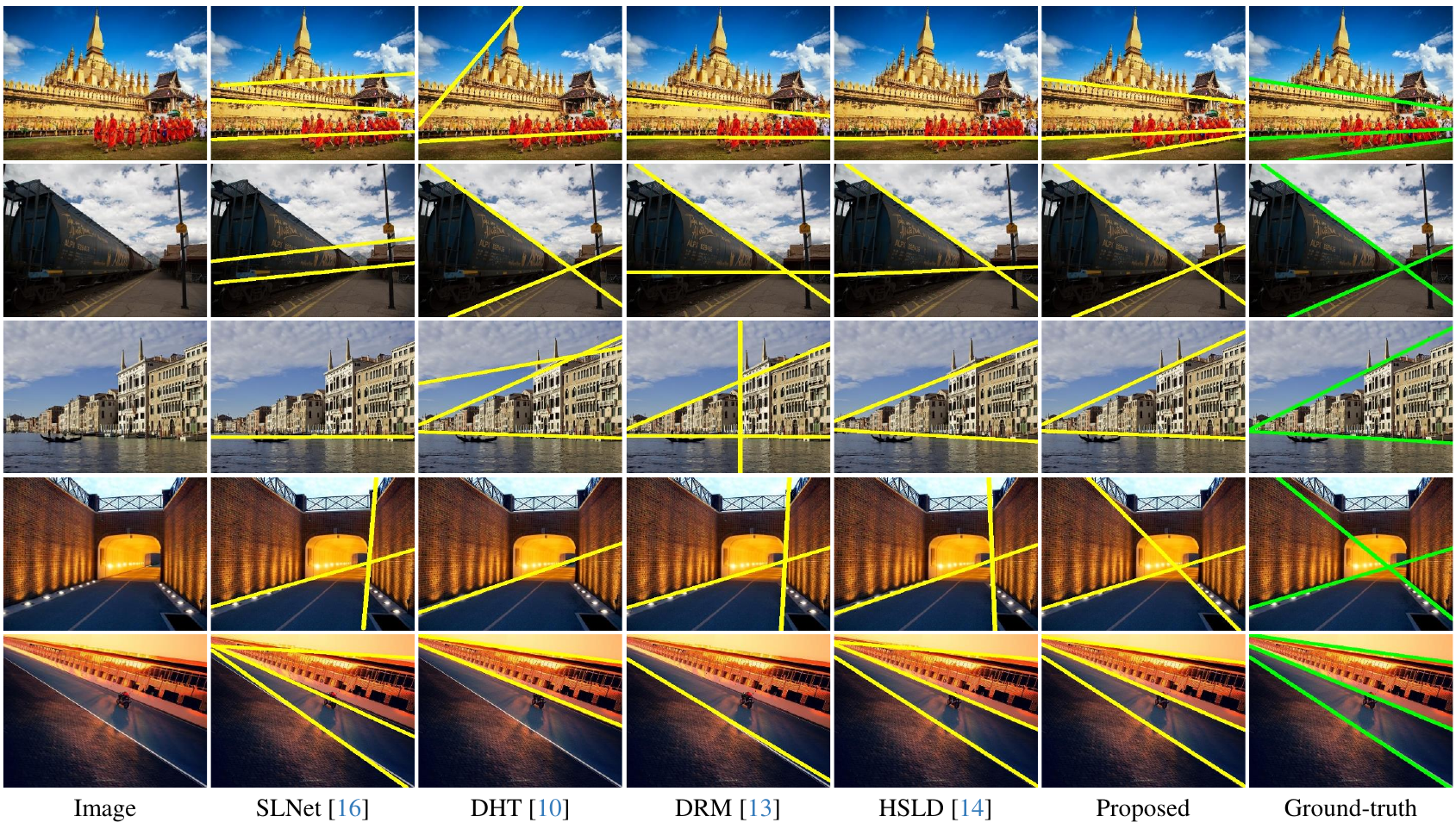}
  \vspace{-0.3cm}
  \caption{Comparison of semantic line detection results on the \textit{Diagonal} class in the CDL dataset.}
  \label{fig:CDL_Diagonal}
\vspace{-0.4cm}
\end{figure}

\begin{figure}[h]
  \centering
  \includegraphics[width=0.98\linewidth]{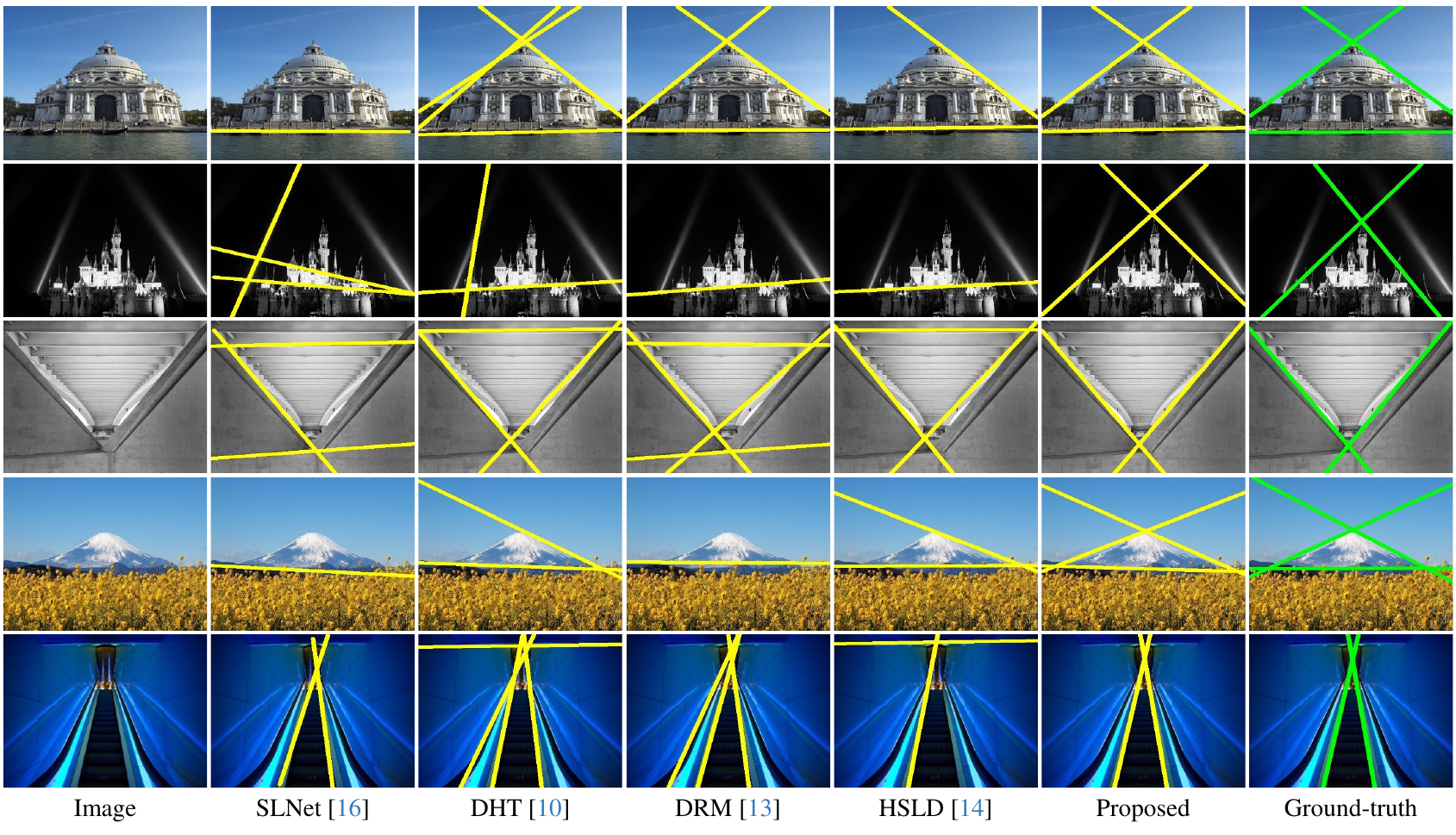}
  \vspace{-0.3cm}
  \caption{Comparison of semantic line detection results on the \textit{Triangle} class in the CDL dataset.}
  \label{fig:CDL_Triangle}
\end{figure}

\clearpage

\begin{figure}[h]
  \centering
  \includegraphics[width=0.98\linewidth]{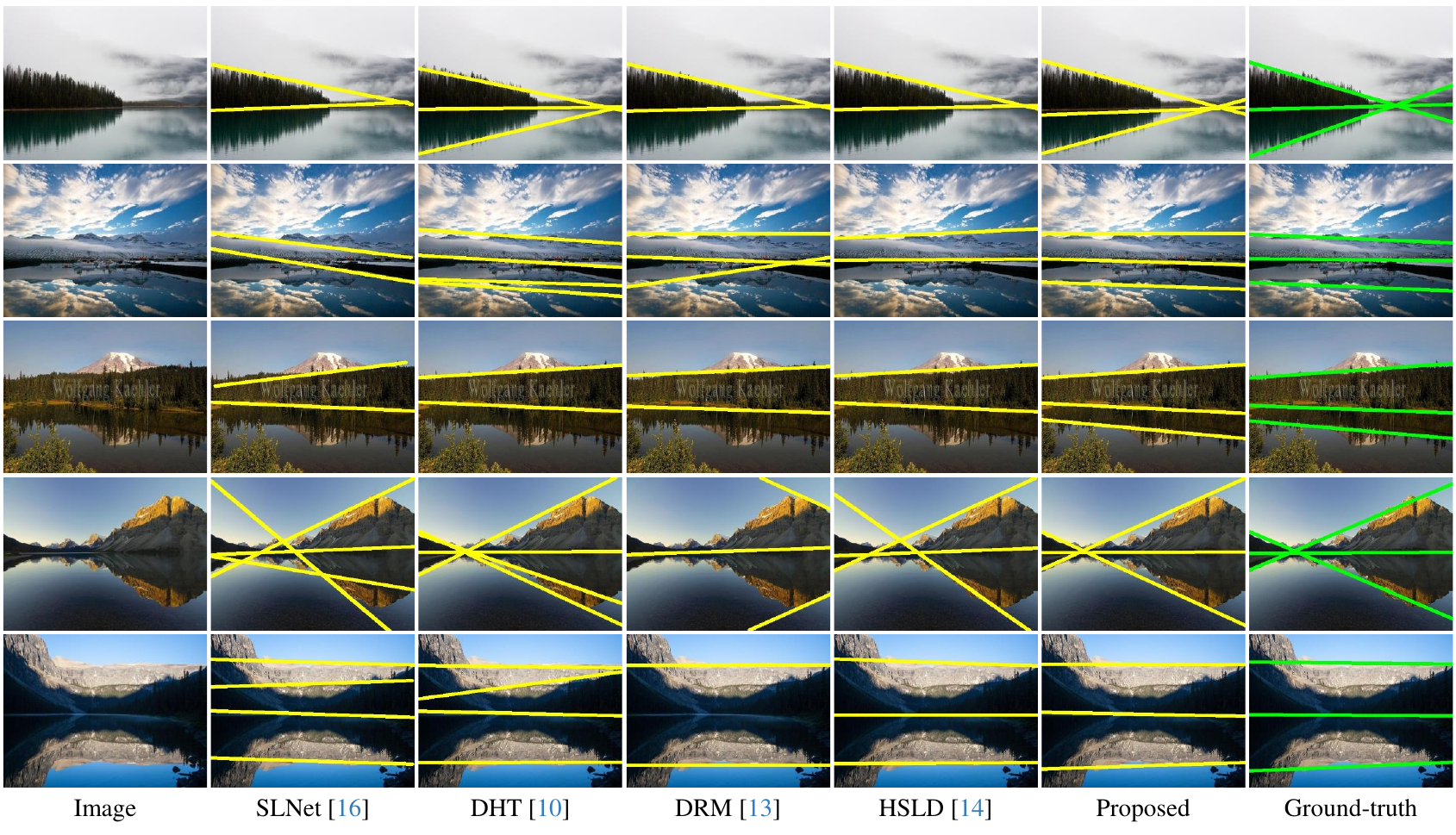}
  \vspace{-0.3cm}
  \caption{Comparison of semantic line detection results on the \textit{Symmetric} class in the CDL dataset.}
  \label{fig:CDL_Symmetric}
\vspace{-0.4cm}
\end{figure}

\begin{figure}[h]
  \centering
  \includegraphics[width=0.98\linewidth]{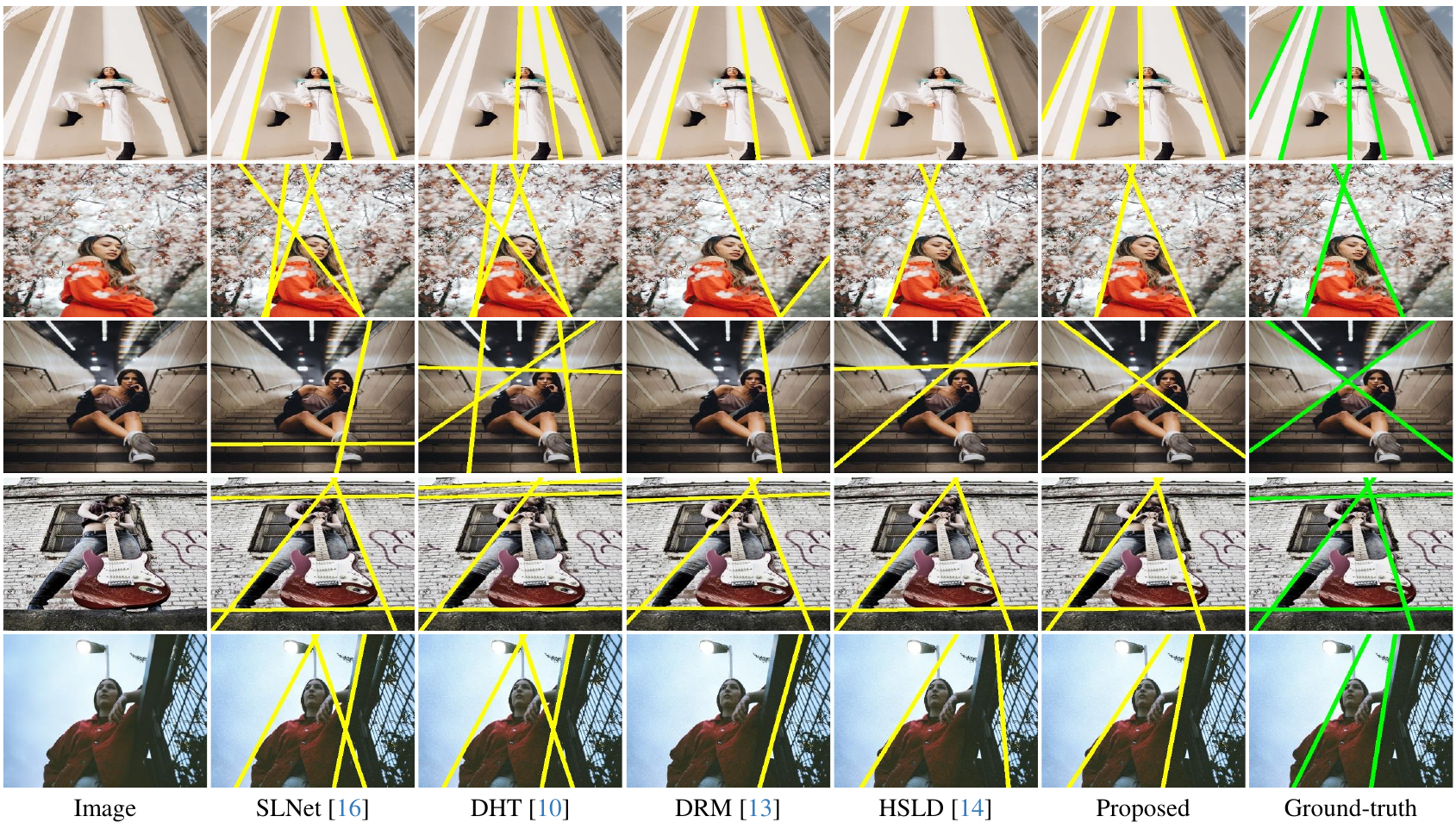}
  \vspace{-0.3cm}
  \caption{Comparison of semantic line detection results on the \textit{Low} class in the CDL dataset.}
  \label{fig:CDL_Low}
\end{figure}

\clearpage

\section{Applications}
\subsection{Dominant vanishing point detection}
We apply the proposed SLCD to detect dominant vanishing points by identifying the vanishing lines. We declare the intersecting point of two lines as a vanishing point. We assess the proposed SLCD on the AVA landscape dataset \cite{zhou2017detecting} which contains 2,000 training and 275 test landscape images. Figure~\ref{fig:VPD} shows more results of vanishing point detection.

\begin{figure}[h]
  \centering
  \includegraphics[width=0.98\linewidth]{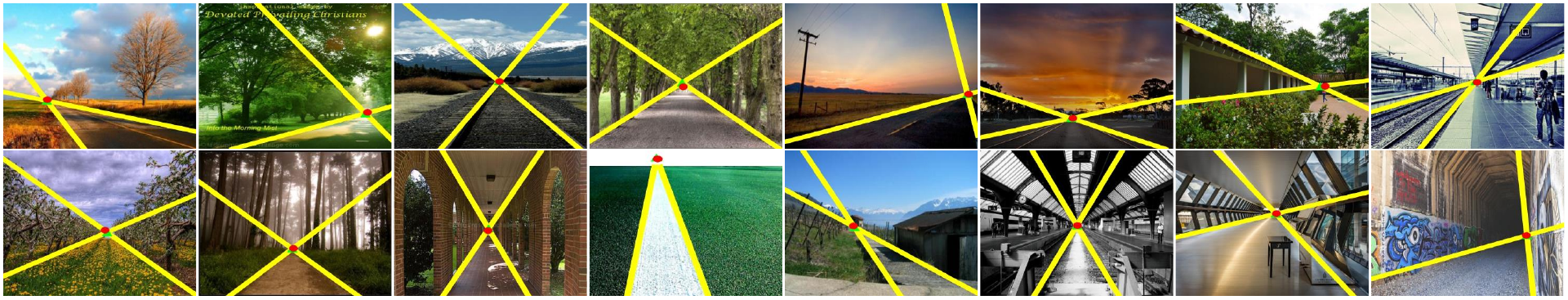}
  \vspace{-0.3cm}
  \caption{More results of vanishing point detection.}
  \label{fig:VPD}
\end{figure}

\clearpage

\subsection{Reflection symmetry axis detection}

We test the proposed SLCD on three datasets: ICCV \cite{funk2017}, NYU \cite{cicconet2017finding}, and SYM\_Hard \cite{jin2020semantic}. ICCV provides 100 training and 96 test images, NYU contains 176 test images, and SYM\_Hard consists of 45 test images. In these datasets, each images contains a single reflection symmetry axis. We trained the model on the ICCV and tested it on all datsets. Figure~\ref{fig:SYM} shows more results of vanishing point detection. In the top row, the ground-truth and predicted axes are depicted by dashed red and solid yellow lines, respectively. The membership maps are also visualized in the bottom row.

\begin{figure}[h]
  \centering
  \includegraphics[width=0.98\linewidth]{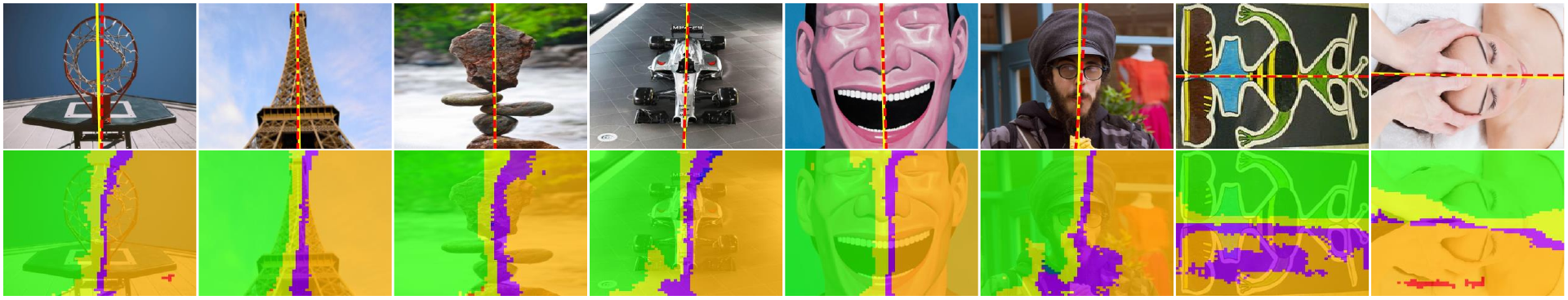}
  \vspace{-0.3cm}
  \caption{More results of symmetric axis detection.}
  \label{fig:SYM}
\end{figure}

\subsection{Composition-based image retrieval}
We test the proposed SLCD on Oxford 5k and Paris 6k dataset \cite{radenovic2018revisiting}, containing photographs of landmarks and local attractions in two places. Since it contains indoor and outdoor images mainly, we use the network trained on NKL dataset. We first detect semantic lines for all the images, while storing the positional feature map $P$. Then, we filter out images whose the composition scores are lower than a threshold $0.75$ since a low score indicates a low quality image with inharmonious composition, as shown in Figure~\ref{fig:Sorting}. Then, for a randomly selected query image, we compute the $\ell_2$-distances between the positional feature maps $P$ of the query and the remaining images. We determine the images with the smallest distances as retrieval results. Figure~\ref{fig:RTR} shows more results of composition-based retrieval.

\begin{figure}[h]
  \centering
  \includegraphics[width=0.98\linewidth]{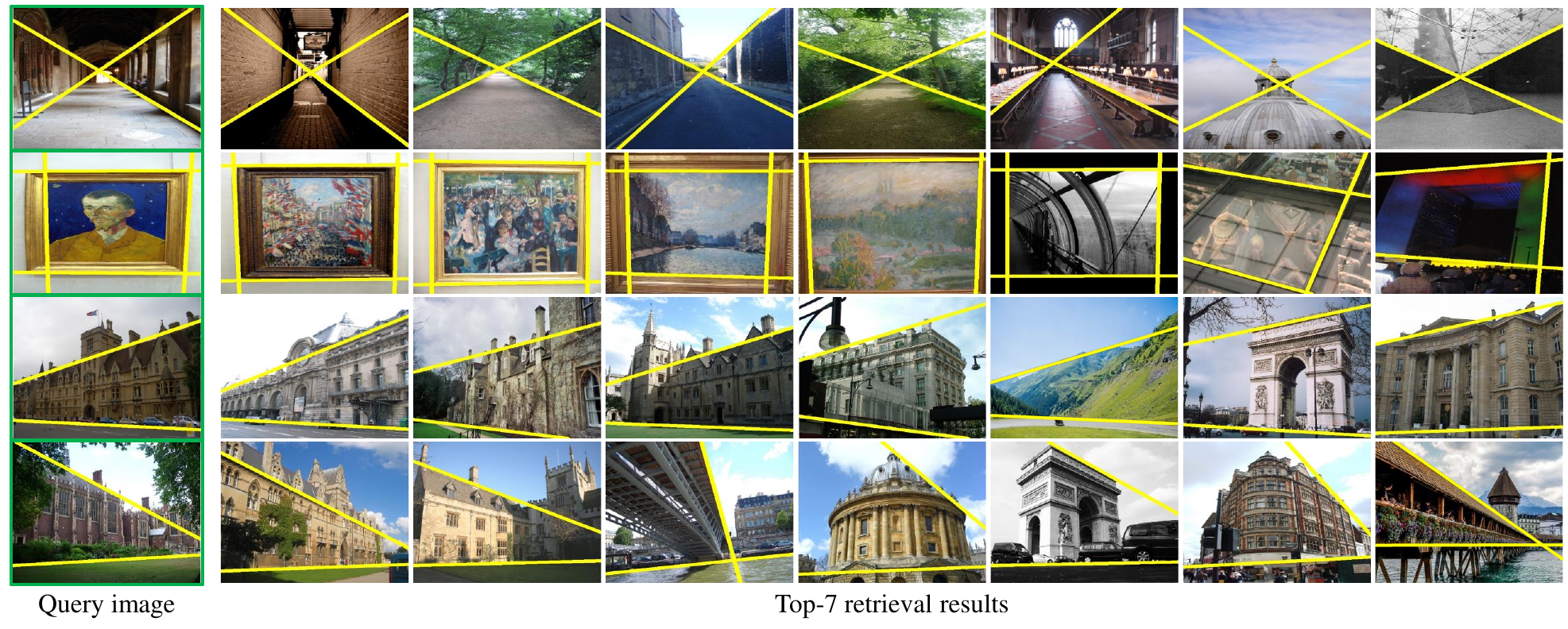}
  \vspace{-0.3cm}
  \caption{More results of composition-based retrieval.}
  \label{fig:RTR}
\end{figure}

\clearpage
\subsection{Road lane detection}
\begin{figure}[h]
  \centering
  \includegraphics[width=0.98\linewidth]{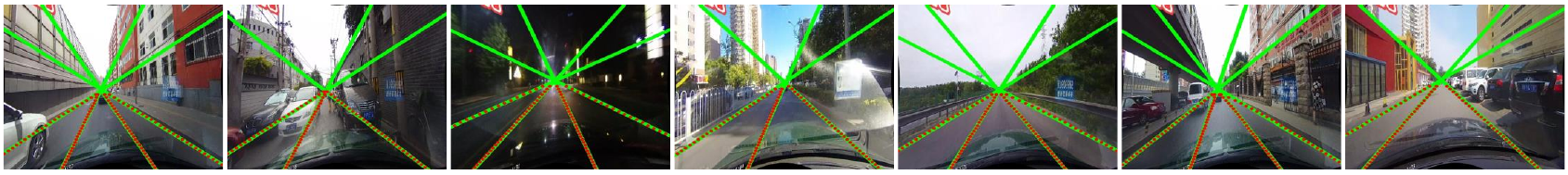}
  \vspace{-0.3cm}
  \caption{Examples of original lane points and ground-truth semantic lines.}
  \label{fig:CULane_GT}
\end{figure}

We apply the proposed SLCD to the lane detection task, by employing the CULane dataset \cite{pan2018spatial}. CULane is a road lane dataset in which lanes are annotated with 2D coordinates. To obtain ground-truth semantic lines corresponding to the lanes in an image, we declare the most overlapping line with each lane as a semantic line. Figure~\ref{fig:CULane_GT} shows some examples of original lane points and ground-truth semantic lines, depicted by red dots and green lines, respectively.

\begin{figure}[h]
  \centering
  \includegraphics[width=0.98\linewidth]{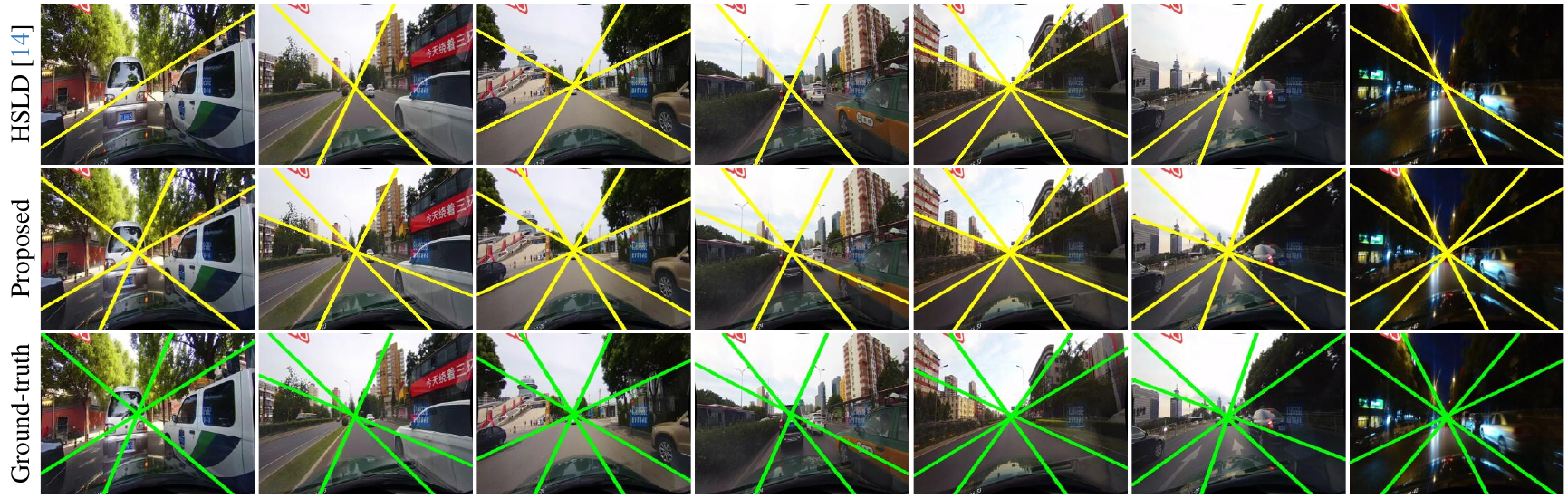}
  \vspace{-0.3cm}
  \caption{Comparison of semantic line detection results on the CULane dataset.}
  \label{fig:CULane_results}
  \vspace{-0.5cm}
\end{figure}

We train SLCD and compare it with the second-best line detector HSLD \cite{jin2021harmonious}. Figure~\ref{fig:CULane_results} shows some lane detection results on the test set. SLCD detects semantic lines more reliably in road scenes than HSLD does, even though some lines are implied due to occlusion or poor illumination. Table \ref{table:RoadLane} compares the HIoU and AUC results. We see that SLCD outperforms HSLD meaningfully.

\begin{table*}[h]\centering
    \renewcommand{\arraystretch}{0.9}
    \caption
    {
        Comparison of the HIoU and AUC results on the CULane dataset.
    }
    \vspace*{-0.15cm}
    \footnotesize
    \begin{tabular}[t]{+L{2.2cm}^C{1.3cm}^C{1.3cm}^C{1.3cm}^C{1.3cm}}
    \toprule
                & AUC\_P & AUC\_R & AUC\_F & HIoU \\
    \midrule
    HSLD \cite{jin2021harmonious}   & 92.38 & 80.11 & 85.80 & 74.14  \\
    Proposed                        & 92.91 & 84.86 & 88.55 & 76.97  \\
    \bottomrule
    \end{tabular}
    \label{table:RoadLane}
    \vspace*{-0.1cm}
\end{table*}

\end{appendices}

\end{document}